\documentclass[preprint,12pt]{elsarticle}

\usepackage[a4paper, left=2cm, right=2cm, top=2cm, bottom=2cm]{geometry}

\usepackage{amssymb}
\usepackage{amsmath}
\usepackage{amsfonts}
\usepackage{multirow}

\usepackage{algorithm}     
\usepackage{algpseudocode} 
\usepackage{amsmath, amsfonts}
\usepackage[cmyk,table]{xcolor}
\usepackage{booktabs,longtable,multirow,makecell}
\usepackage{array}          
\newcolumntype{C}[1]{>{\centering\arraybackslash}p{#1}}        
 
\usepackage{adjustbox}
\usepackage{hyperref}

\journal{Information Fusion}

\begin{document}

\begin{frontmatter}

\title{\textbf{CX-Mind}: A Pioneering Multimodal Large Language Model for Interleaved Reasoning in Chest X-ray via Curriculum-Guided Reinforcement Learning\tnotemark[*]}

\author[a,b,c]{Wenjie Li\fnmark[1]}
\author[b]{Yujie Zhang\fnmark[1]}
\author[d]{Haoran Sun\fnmark[1]}
\author[e]{Yueqi Li}
\author[b,f]{Fanrui Zhang}
\author[g]{Mengzhe Xu}
\author[h]{Victoria Borja Clausich}              
\author[i]{Sade Mellin}                          
\author[a,c]{Renhao Yang}
\author[b,j]{Chenrun Wang}                          
\author[k,l]{Jethro Zih-Shuo Wang}                       
\author[c]{Shiyi Yao}
\author[c]{Gen Li}
\author[c,m]{Yidong Xu}                          
\author[c]{Hanyu Wang}
\author[c]{Yilin Huang}
\author[c]{Angela Lin Wang}
\author[c]{Chen Shi}
\author[c]{Yin Zhang}
\author[c]{Jianan Guo}
\author[c]{Luqi Yang}
\author[c]{Renxuan Li}
\author[c]{Yang Xu}
\author[g]{Jiawei Liu}
\author[n]{Yao Zhang}
\author[d]{Lei Liu}
\author[o]{Carlos Gutiérrez SanRomán}             
\author[a,c]{Lei Wang\corref{cor1}}

\fntext[1]{Equal contributions.}
\cortext[cor1]{Corresponding author at: College of Health Science and Technology, Shanghai Jiao Tong University School of Medicine, Shanghai, China.\\
E-mail: ray\_wangs@hotmail.com (L.\,Wang)}
\tnotetext[*]{The source code of the paper will be released at \href{https://github.com/WenjieLisjtu/CX-Mind}{https://github.com/WenjieLisjtu/CX-Mind}.}

\affiliation[a]{organization={College of Health Science and Technology, Shanghai Jiao Tong University School of Medicine},
               city={Shanghai},
               country={China}}

\affiliation[b]{organization={Shanghai Innovation Institute},
               city={Shanghai},
               country={China}}

\affiliation[c]{organization={Clinical Center for Sports Medicine, Department of Orthopaedics, Ruijin Hospital, Shanghai Jiao Tong University School of Medicine},
               city={Shanghai},
               country={China}}

\affiliation[d]{organization={School of Basic Medical Sciences, Intelligent Medicine Institute, Fudan University},
               city={Shanghai},
               country={China}}

\affiliation[e]{organization={Department of Hematology, The First Affiliated Hospital, College of Medicine, Zhejiang University},
               city={Hangzhou},
               country={China}}

\affiliation[f]{organization={MoE Key Laboratory of Brain-Inspired Intelligent Perception and Cognition, University of Science and Technology of China},
               city={Hefei},
               country={China}}

\affiliation[g]{organization={Department of Public Health and Primary Care, University of Cambridge},
               city={Cambridge},
               country={United Kingdom}}

\affiliation[h]{organization={Department of Medicine, Faculty of Health Sciences, Universidad CEU Cardenal Herrera},
               city={Valencia},
               country={Spain}}

\affiliation[i]{organization={Faculty of Medicine, University of Helsinki},
               city={Helsinki},
               country={Finland}}

\affiliation[j]{organization={X-LANCE Lab, School of Computer Science, Shanghai Jiao Tong University},
               city={Shanghai},
               country={China}}

\affiliation[k]{organization={Department of Hepatobiliary Surgery, National Cancer Center / National Clinical Research Center for Cancer / Cancer Hospital, Chinese Academy of Medical Sciences and Peking Union Medical College},
               city={Beijing},
               country={China}}

\affiliation[l]{organization={Department of Surgery, The Ohio State University Wexner Medical Center, The James Comprehensive Cancer Center},
               city={Columbus},
               country={United States}}

\affiliation[m]{organization={Ningbo Institute of Technology, Beihang University},
               city={Ningbo},
               country={China}}

\affiliation[n]{organization={Department of Mechanical Engineering, University College London},
               city={London},
               country={United Kingdom}}

\affiliation[o]{organization={Department of Pediatric Surgery (Servicio Cirugía Pediátrica), Hospital Universitario y Politécnico La Fe},
               city={Valencia},
               country={Spain}}

\begin{abstract}
Chest X-ray (CXR) imaging is one of the most widely used diagnostic modalities in clinical practice, encompassing a broad spectrum of diagnostic tasks. Recent advancements have seen the extensive application of reasoning-based multimodal large language models (MLLMs) in medical imaging to enhance diagnostic efficiency and interpretability. However, existing multimodal models predominantly rely on "one-time" diagnostic approaches, lacking verifiable supervision of the reasoning process. This leads to challenges in multi-task CXR diagnosis, including lengthy reasoning, sparse rewards, and frequent hallucinations. To address these issues, we propose \textbf{CX-Mind}, the first generative model to achieve interleaved "think-answer" reasoning for CXR tasks, driven by curriculum-based reinforcement learning and verifiable process rewards (CuRL‑VPR). Specifically, we constructed an instruction-tuning dataset, CX-Set, comprising 708,473 images and 2,619,148 samples, and generated 42,828 high-quality interleaved reasoning data points supervised by clinical reports. Optimization was conducted in two stages under the Group Relative Policy Optimization framework: initially stabilizing basic reasoning with closed-domain tasks, followed by transfer to open-domain diagnostics, incorporating rule-based conditional process rewards to bypass the need for pretrained reward models. Extensive experimental results demonstrate that \textbf{CX-Mind} significantly outperforms existing medical and general-domain MLLMs in visual understanding, text generation, and spatiotemporal alignment, achieving an average performance improvement of 25.1\% over comparable CXR-specific models. On real-world clinical dataset (Rui-CXR), \textbf{CX-Mind} achieves a mean recall@1 across 14 diseases that substantially surpasses the second-best results, with multi-center expert evaluations further confirming its clinical utility across multiple dimensions. \textbf{CX-Mind} establishes a new paradigm for constructing interpretable, and high-performing medical MLLMs.

\end{abstract}



\begin{keyword}
Medical Reasoning; 
Multimodal Large Language Models; 
Chest X-ray; 
Reinforcement Learning; 
Curriculum Learning

\end{keyword}

\end{frontmatter}

\section{Introduction}
\label{sec1}

\begin{figure}[tbp]
    \centering 
    \includegraphics[width=0.95\textwidth]{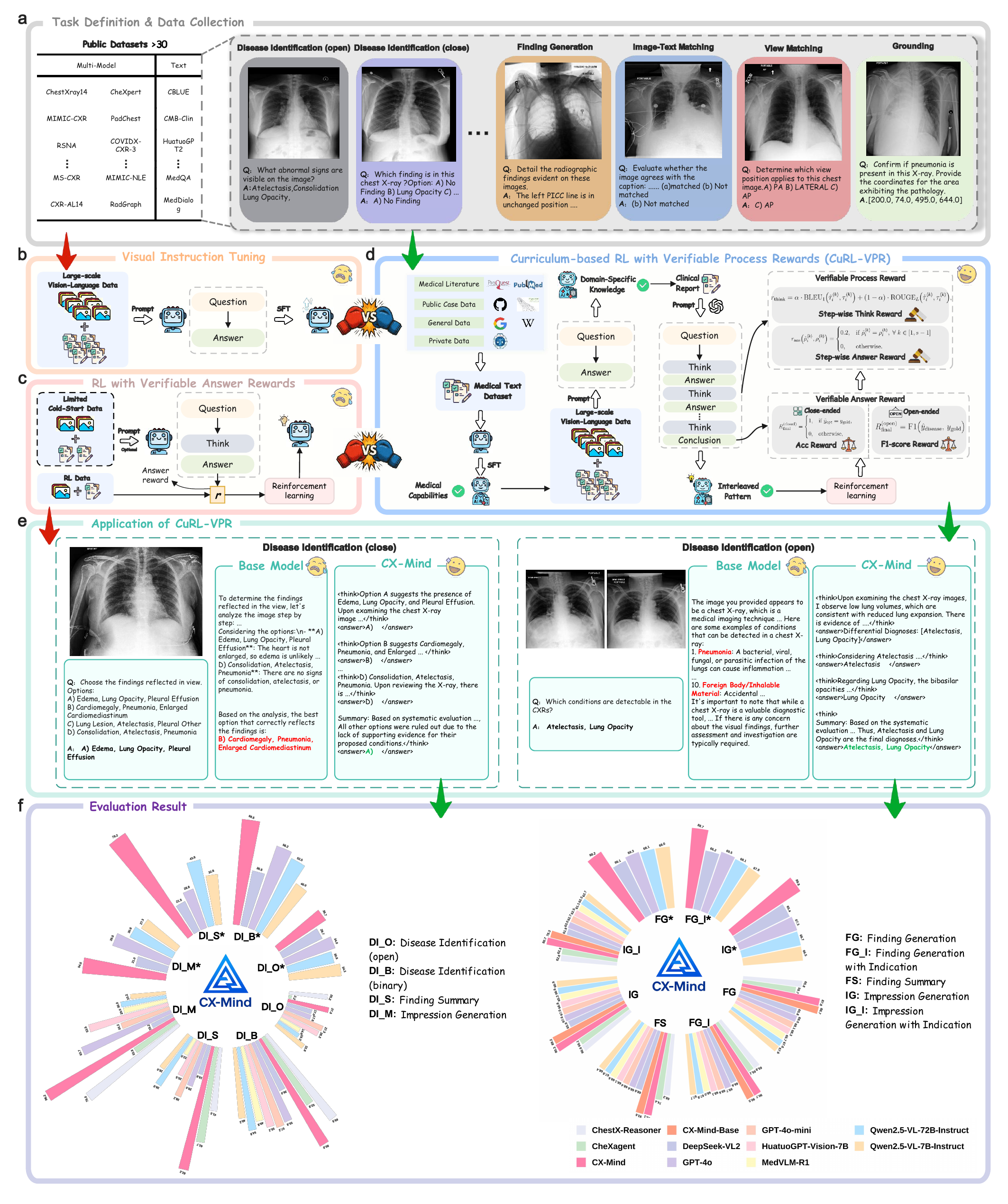} 
    \caption{\textbf{CX-Mind: An MLLM for interleaved reasoning in Chest X-ray.}
(a) Task Definition and Data Collection: A broad suite of chest X-ray interpretation tasks, with more than 30 single-model and multimodal public datasets collected.
(b) Visual Instruction Tuning: Fine-tuning with large-scale instruction data.
(c) RL with Verifiable Answer Rewards: Reinforcement learning driven solely by outcome (answer) rewards.
(d) CuRL-VPR framework: A curriculum-style RL fine-tuning paradigm that mines real reports to achieve process supervision.
(e) Evaluation Result: Summary of CX-Mind versus numerous baselines, reporting Accuracy for visual understanding and BERTScore for text generation, and * denotes out-of-domain testing.}

    \label{intro} 
\end{figure}

In the past year alone, the global landscape of large language models (LLMs) has undergone a "reasoning revolution": OpenAI-o1 pioneered the introduction of extended Chain-of-Thought (CoT) pathways \citep{jaech2024openai}, while models such as DeepSeek-R1 and Gemini 2.5 Pro have continued to push the boundaries of reasoning capabilities \citep{doshi2025gemini,guo2025deepseek}. Empirical evidence has demonstrated that providing models with sufficient "reasoning space" enables them to exhibit near-expert-level proficiency in highly constrained tasks, such as mathematical proofs and structured programming. This momentum has rapidly transcended textual domains, extending into multimodal scenarios. When visual and linguistic signals are deeply integrated and mapped to a unified semantic space, CoT reasoning significantly amplifies model capabilities, consistently consistently reaching new heights in cross-modal inference tasks \citep{wang2025multimodal}.

In the medical domain, particularly in the high-frequency and task-diverse context of chest X-ray imaging, the development of user-friendly and interpretable AI assistants is imperative. Clinicians expect iterative feedback tracing evidence chains to conclusions, rather than opaque, one-shot answers. Only through reviewable and queryable intermediate conclusions can clinical personnel promptly identify and correct errors, ensuring safety and compliance \citep{holzinger2022information,ali2023explainable}. As shown in Figure \ref{intro}(b), some studies have attempted to simulate "doctor-like reasoning" by leveraging large-scale chest X-ray report instruction fine-tuning or incorporating CoT prompts and reasoning process text in medical visual question answering (VQA) \citep{chen2024vision,fallahpour2025medrax}. As depicted in Figure \ref{intro}(c), other researches have employed reinforcement learning (RL) to enhance reasoning capabilities, designing rewards at the answer level or relying on additionally trained reward models \citep{lai2025med,pan2025medvlm}. These efforts have significantly improved model performance in medical tasks, validating the value of CoT reasoning and RL. However, existing medical reasoning multimodal large language models (MLLMs) predominantly adhere to a "one-shot judgment" paradigm, delivering final answers after protracted reasoning processes. This approach lacks verifiable process supervision, making it challenging for clinicians to intervene and correct errors in a timely manner \citep{pan2025medvlm,sharma2024cxr}. Furthermore, reliance on reward signals tied to final outcomes exacerbates the risk of hallucinations \citep{albahri2023systematic,xu2025knowledge}. Consequently, there is an urgent need in the chest X-ray domain for a novel paradigm that supports interleaved reasoning while mitigating hallucinations.

To address these challenges, we propose \textbf{\textbf{CX-Mind}}: a reasoning-focused multimodal large language model tailored for chest X-ray diagnostics, with "think–answer" interleaved reasoning as its fundamental interaction unit. This approach ensures interpretability and reviewability while significantly enhancing multi-task diagnostic performance. Specifically, we adopt curriculum-based reinforcement learning with verifiable process rewards (CuRL-VPR) as the core optimization strategy, with the overall workflow illustrated in Figure \ref{intro}.

We first systematically outline three essential capabilities required for a chest X-ray reasoning model: visual understanding, text generation, and spatiotemporal alignment. As illustrated in Figure \ref{intro}(a), inspired by \citep{chen2024vision}, we decompose each capability into task subsets with clearly defined input-output structures. We collected over 30 publicly available datasets and, based on these, utilized DeepSeek-V3 \citep{liu2024deepseek} and GPT-4o \citep{hurst2024gpt} models to automatically construct a large-scale instruction dataset, CX-Set, comprising 708,473 images and 2,619,148 samples, along with 42,828 high-quality interleaved reasoning samples generated under supervision from real-world reports, providing robust data support for the interleaved reasoning paradigm.

Regarding training, as shown in Figure \ref{intro}(d), we adopt a four-stage curriculum design. First, we fine-tune the language model component of our multimodal model using clinical corpora, enabling it to master specialized medical terminology and reasoning patterns. Second, large-scale chest X-ray instruction fine-tuning injects vision-language knowledge and establishes robust semantic alignment for imaging data. Inspired by DeepSeek-R1 \citep{guo2025deepseek}, we leverage RL to stimulate the model’s reasoning capability through a two-stage training approach: the third stage uses a hybrid of answer-only and interleaved reasoning samples for cold-start supervision, allowing the model to learn the interleaved output format and providing a stable starting point for subsequent policy optimization. The fourth stage, conducted within the Group Relative Policy Optimization (GRPO) framework, similarly employs a curriculum-based approach for RL. It begins with close-ended tasks to construct stable and verifiable reward signals, then transitions to open-ended diagnostics to achieve higher-level free reasoning capabilities \citep{rui2025improving}. Unlike traditional methods that rely solely on final answer rewards, our proposed verifiable process reward mechanism provides fine-grained feedback after each think–answer pair and employs conditional incentives, mitigating the credit assignment problem.

In the evaluation phase, we constructed a comprehensive benchmark covering open-source datasets such as MIMIC-CXR, CheXpert, and MS-CXR, as well as proprietary data from Rui-CXR. This benchmark includes over 30K VQA samples, assessing multiple tasks under each capability domain. As shown in Figure \ref{intro}(f), experimental results demonstrate that \textbf{CX-Mind} significantly outperforms existing general-purpose and medical-specialized MLLMs across all three capability domains, achieving an average performance improvement of 25.1\% compared to chest X-ray-specific models. More importantly, the interleaved output showcased in Figure \ref{intro}(e) avoids lengthy and cumbersome reasoning chains generated by traditional CoT approaches. Furthermore, we conducted comprehensive ablation studies and large-scale validation on the real-world clinical dataset Rui-CXR, accompanied by multidimensional evaluations from multiple clinical experts.

In summary, the main contributions of this paper can be outlined in six key points:
\begin{itemize}
\item We introduce a large-scale chest X-ray VQA dataset, CX-Set, encompassing 23 related datasets and 13 distinct tasks, with a total of over 2 million data entries. Additionally, we construct over 40K high-quality interleaved CoT data based on supervised patterns derived from real-world reports.
\item We introduce a novel stage-wise training strategy guided by curriculum learning, which progressively injects chest X-ray-related medical knowledge and enhances the reasoning capabilities of MLLMs.
\item We establish the first interleaved reasoning paradigm for multimodal medical applications, which enhances the interpretability of reasoning models and facilitates effective communication during human interaction.
\item We design an effective rule-based process reward method to cultivate the predefined interleaved reasoning capabilities of MLLMs, with targeted training for both close-ended and open-ended questions.
\item We demonstrate that \textbf{CX-Mind} achieves competitive performance compared to previous state-of-the-art (SOTA) medical reasoning MLLMs through extensive experiments across multiple benchmarks.
\item We further conduct external validation using a collected real-world clinical dataset, Rui-CXR, and recruit three clinical experts to comprehensively evaluate \textbf{CX-Mind} across five distinct metrics to demonstrate its clinical utility.
\end{itemize}

\section{Related Works}

\subsection{X-ray Foundation Model}
Recent advancements in foundation models have revolutionized chest X-ray analysis, driving significant progress in image generation, disease diagnosis and radiology report generation. These innovations leverage cutting-edge techniques such as domain-adapted pretraining and self-supervised learning to address persistent challenges like data scarcity and model generalizability.
For instance, Chambon et al. \citep{bluethgen2024vision} introduced a vision-language foundation model that generates highly realistic synthetic chest X-ray images by fine-tuning latent diffusion models on public datasets.
Transitioning to disease diagnosis, Wang et al. \citep{ma2025fully} developed Ark+, an open-source model pretrained on diverse datasets with heterogeneous expert annotations. Ark+ achieves high diagnostic accuracy across multiple chest diseases while supporting privacy-preserving federated learning, making it a practical tool for clinical use. Complementing this, Zhang et al. \citep{yao2024eva} proposed EVA-X, which uses self-supervised learning to extract semantic and geometric features from unlabeled images. By reducing reliance on annotated data, EVA-X excels in over 11 detection tasks, advancing the field’s ability to address data scarcity.
In the realm of radiology report generation, Chen et al. \citep{chen2024chexagent} introduced CheXagent, trained on the large-scale CheXinstruct dataset. CheXagent significantly reduces clinicians’ report drafting time, as validated by the CheXbench benchmark, enhancing clinical efficiency. Similarly, Lee et al. \citep{lee2024comparative} compared M4CXR, a domain-specific MLLM, with ChatGPT-4o, showing that M4CXR outperforms in diagnostic accuracy and report generation due to its tailored adaptation to chest X-ray tasks.
However, existing foundation models have not applied advanced techniques for enhancing reasoning capabilities to chest X-ray tasks. \textbf{CX-Mind}, through a multi-stage curriculum learning strategy, not only improves its ability to handle various fundamental clinical tasks but also enhances its reasoning capabilities for addressing complex problems.

\subsection{Reasoning in Medical MLLMs}
Advancements in LLMs have enhanced vision-language models, enabling them to address complex visual reasoning tasks that demand robust visual perception and advanced cognitive capabilities. Recent studies have optimized visual encoding strategies to improve the quality of visual tokens. Techniques such as prompt tuning, supervised fine-tuning (SFT), and RL have proven critical in enhancing the reasoning capabilities of MLLMs.
In medical domains, where interpretability is paramount, models like HuatuoGPT-o1 \citep{chen2024huatuogpt} and Baichuan-M1 \citep{wang2025baichuan} emphasize transparent reasoning paths to tackle medical challenges. However, integrating multimodal medical data for reasoning in MLLMs remains in its early stages. Well-crafted prompts simulating doctor-like reasoning processes show promise but are limited in applicability. Leveraging DeepSeek-R1’s success with RL, GRPO employs rule-based reward functions to improve adaptability across diverse medical tasks. For example, Med-R1 \citep{lai2025med}, utilizing GRPO, demonstrates superior performance across eight medical image modalities and five question types, enhancing generalizability and reliability. Similarly, MedVLM-R1 \citep{pan2025medvlm} adopts an RL framework to promote interpretable reasoning paths without reliance on reference reasoning, mitigating overfitting issues associated with SFT.
Despite these advancements, current models have yet to apply specialized medical reasoning to diagnostic tasks involving chest X-rays. Our proposed \textbf{CX-Mind} addresses this gap, enabling enhanced interaction with clinicians through interleaved reasoning processes.

\subsection{RL for MLLMs Reasoning}

RL has become a powerful approach for enhancing the reasoning capabilities of MLLMs, which integrate text and image data to perform tasks like visual question answering, image captioning, and clinical decision support. RL enables MLLMs to refine decision-making by optimizing reward functions, improving logical coherence and robustness over traditional SFT, which often struggles with overfitting and limited generalization.
In general applications, RL techniques like GRPO guide MLLMs to focus on relevant visual and textual features, enhancing performance in complex reasoning tasks. 
In medical contexts, RL’s impact is particularly significant, as it supports interpretable and trustworthy reasoning across diverse data sources, such as medical images (e.g., X-rays, MRIs) and clinical notes. For instance, Lai et al. introduced Med-R1 \citep{lai2025med}, which leverages GRPO to achieve a 29.94\% accuracy improvement over its base model (Qwen2-VL-2B) across eight medical imaging modalities, outperforming larger models and enhancing cross-task generalization by 32.06\%. Similarly, Pan et al. proposed MedVLM-R1 \citep{pan2025medvlm}, which uses RL to foster interpretable reasoning paths without reference reasoning, improving accuracy from 55.11\% to 78.22\% on MRI, CT, and X-ray benchmarks with minimal training data. Other works, such as Patho-R1 \citep{zhang2025patho} and studies on radiology report generation, further demonstrate RL’s versatility in enhancing specialized medical reasoning.
Although these methods integrate RL with medical tasks, they rely solely on the final answer as the basis for reward computation. In contrast, we propose a novel rule-based reward mechanism that encompasses the reasoning process, intermediate results, and the final answer. This approach is designed to train \textbf{CX-Mind}, thereby reducing hallucinations and enhancing reasoning accuracy.

\section{Methodology}
\label{method}

\begin{figure}[tbp]
    \centering 
    \includegraphics[width=1\textwidth]{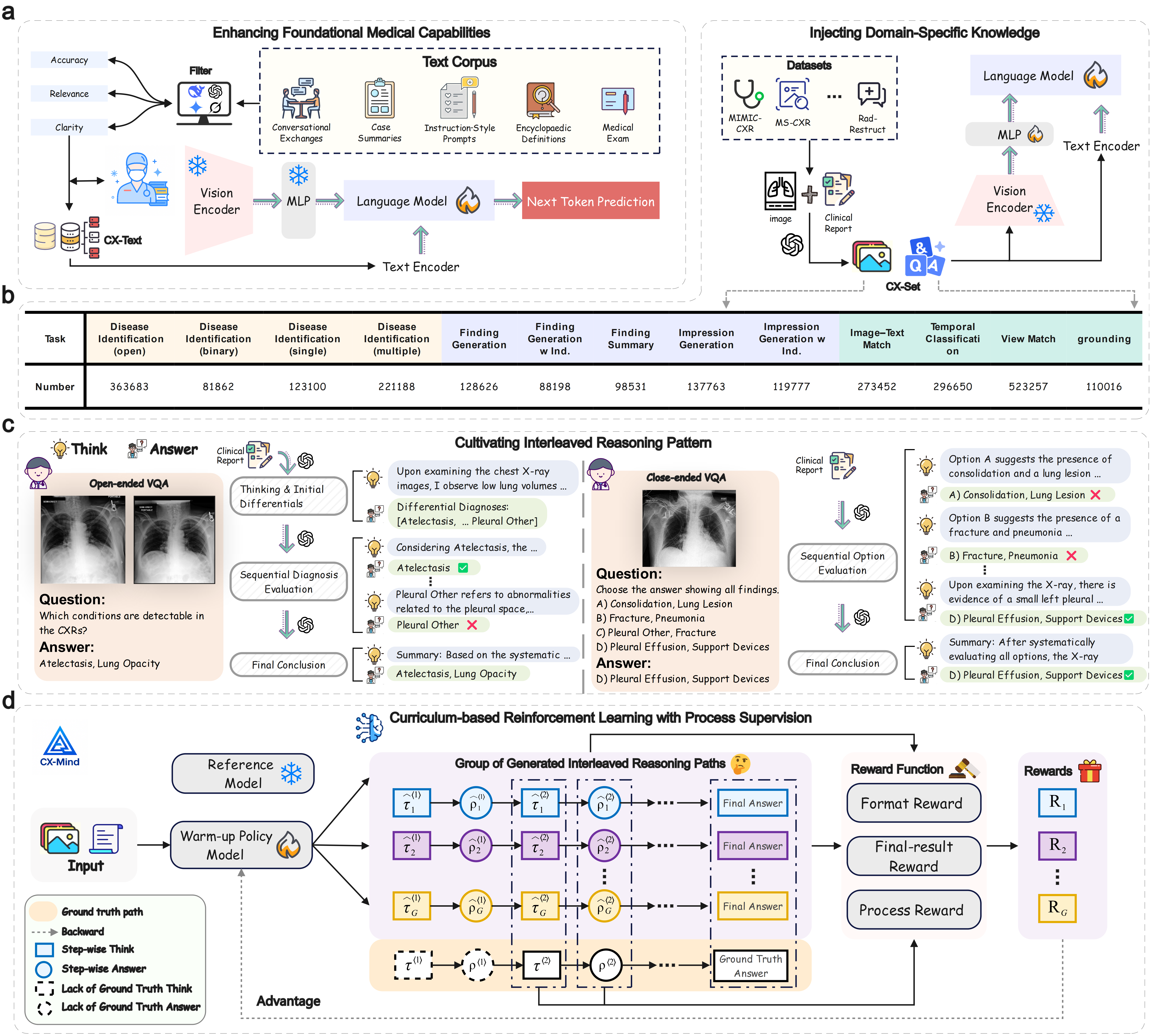} 
    \caption{\textbf{CX-Mind training pipeline.}
(a) Enhancing Foundational Medical Capabilities: We collect public text corpora and filter them with LLM-based scoring to obtain the CX-Text, then warm up the language-model component.
(b) Injecting Domain-Specific Knowledge: We curate and construct over two million QA pairs to cultivate professional X-ray interpretation capability.
(c) Cultivating the Interleaved Reasoning Pattern: Based on real medical reports, we prompt GPT-4o to build different forms of interleaved reasoning chains for close-ended and open-ended questions, thereby cold-starting the model.
(d) RL with Process Supervision: The entire optimization is conducted with GRPO. The warm-up policy model $\pi_\theta$ generates interleaved reasoning paths. Beyond outcome and format rewards, we further reward key reasoning steps. }

    \label{fig:method} 
\end{figure}

As described in Section \ref{sec1}, traditional CoT reasoning employs a "think-then-answer" approach, where the thinking component is often excessively verbose. We illustrate an example of the proposed interleaved diagnostic reasoning in Figure \ref{fig:method}(c). Furthermore, existing multimodal reasoning models only apply verifiable reinforcement learning to the final results, lacking supervision over the reasoning process. To address this, we propose curriculum-based reinforcement learning and verifiable process rewards (CuRL-VPR). Specifically, as shown in Figure \ref{fig:method}, we introduce a fine-tuning paradigm for medical models that integrates SFT and open-ended RL. This paradigm, grounded in a curriculum learning strategy, divides the training process into four stages. To ensure the verifiability and transferability of rewards, we exclusively employ rule-based multidimensional evaluators, mitigating the risk of reward hacking.

In the following, this section will elaborate on the curriculum learning strategy in a stepwise manner: (1) Section \ref{sec3.2} introduces multiple metrics to filter high-quality textual medical data for enhancing foundational medical capabilities, (2) Section \ref{sec3.3} constructs a large-scale instruction-following dataset, CX-Set, for knowledge injection, (3) Section \ref{sec3.4} generates interleaved CoT data for cold-start training, (4) Section \ref{sec3.5} implements curriculum-based reinforcement learning with process supervision.

\subsection{Problem Definition} 

Unlike conventional approaches that rely on a vision–language model (VLM) to render a single‑pass judgement, we introduce an interleaved diagnostic reasoning paradigm in which the model alternates between visual evidence parsing and linguistic inference, performing step‑wise differential refinement to achieve both higher accuracy and stronger interpretability for chest X‑ray analysis. Concretely, we construct a multimodal dataset $D$:
\begin{equation}
D
=\Bigl\{
      \bigl(
        \mathbf{X}_i,\,
        \mathbf{C}_i,\,
        \mathbf{Q}_i,\,
        \mathbf{R}_i,\,
        \mathbf{S}_i,\,
        \mathbf{A}_i
      \bigr)
\Bigr\}_{i=1}^{N},
\end{equation}
where \(N\) is the number of studies, \(\mathbf{X}_i=\{x_{i}^{(1)},\dots,x_{i}^{(K_i)}\}\) are the multi‑view radiographs from the same examination, \(\mathbf{C}_i\) denotes the textual context, \(\mathbf{Q}_i\) contains both open-ended and close-ended questions, \(\mathbf{S}_i\) is the reference report, and
$
    \mathbf{R}_i=
\bigl[(\tau_{i}^{\langle1\rangle},\rho_{i}^{\langle1\rangle}),\dots,
      (\tau_{i}^{\langle T_i\rangle},\rho_{i}^{\langle T_i\rangle})\bigr]
$
is the ground‑truth interleaved reasoning chain whose elements \((\tau_{i}^{\langle s\rangle},\rho_{i}^{\langle s\rangle})\) represent the \(s\)-th thought fragment and its accompanying answer snippet guided by \(\mathbf{S}_i\), and \(\mathbf{A}_i\) records the definitive diagnosis and lesion coordinates.

We fine‑tune an open‑source multimodal large language model \(\pi_{\theta}\) so that, given only the images \(\mathbf{X}\), context \(\mathbf{C}\), and question \(\mathbf{Q}\), it auto‑regressively produces the interleaved sequence:
\begin{equation}
\label{equation2}
\mathcal{Y}\;=\;\underbrace{\tau^{\langle1\rangle}}_{\text{think}}\!
           \oplus\!
           \underbrace{\rho^{\langle1\rangle}}_{\text{answer}}\!
           \oplus\!\;
           \dots 
           \oplus
           \tau^{\langle s\rangle}
           \oplus
           \rho^{\langle s\rangle}
\;=\;
\pi_{\theta} \bigl(\cdot\mid\mathbf{X},\mathbf{C},\mathbf{Q}\bigr),
\end{equation}
where each pair \((\tau^{\langle s\rangle},\rho^{\langle s\rangle})\) is supervised by the corresponding ground‑truth step in \(\mathbf{R}\). The final answer fragment \(\rho^{\langle s\rangle}\) must jointly realise lesion localisation, differential diagnosis, and report generation.

We partition the dataset as:
\begin{equation}
D
=
D_{R}
\cup
D_{A} \;,\;
D_{R}
=
D_{R_o}
\cup
D_{R_c},
\end{equation}
where \(D_{R}\) provides complete reasoning chains \((\mathbf{R},\mathbf{S},\mathbf{A})\) and is further split into \(D_{R_o}\) (open‑ended queries) and \(D_{R_c}\) (close‑ended queries), whereas \(D_{A}\) contains answer‑only samples without \(\mathbf{R}\). The utilisation strategy of these subsets is detailed in subsequent section.

\subsection{Enhancing Foundational Medical Capabilities}
\label{sec3.2}
To equip the model with foundational medical vocabulary and clinical reasoning capabilities, we begin our curriculum with a concise, text-only SFT phase. Recent studies have indicated that such a targeted textual warm-up can significantly facilitate subsequent multimodal optimisation. For instance, HealthGPT reported improved stability and faster convergence following an initial SFT on clinical documents \citep{lin2025healthgpt}, while BiomedGPT observed notable performance improvements in medical vision-language tasks when the vision encoder was kept frozen during an early textual training stage \citep{luo2024biomedgpt}. Motivated by these findings, we restrict updates to the language layers in this stage, maintaining all visual parameters frozen in Figure \ref{fig:method}(a). This ensures that the model internalises domain-specific knowledge before visual features are integrated.

The warm‑up corpus comprises roughly 200K English question–answer (QA) pairs automatically translated from eight public resources. In terms of data types, it spans (i) conversational exchanges, (ii) structured case summaries, (iii) instruction‑style prompts, (iv) encyclopaedic definitions, and (v) medical examination items. Concretely, we draw from CBLUE, CMB‑Clin, HuatuoGPT2, CMExam, CMB‑Exam, MedDialog and MedQA. This mixture exposes the model to a broad range of clinical entities, diagnostic cues, therapeutic options, and general medical reasoning patterns.

Each candidate QA pair is then evaluated by LLMs that serves as an automatic judge. We adopt a five‑point scale on three enriched axes. \textbf{Accuracy} assesses whether the text faithfully reflects medical facts, including symptoms, diagnoses, and treatment plans, and whether it is consistent with recognised clinical practice. \textbf{Relevance} measures how closely the content aligns with radiological diagnostic themes, penalising digressions into unrelated specialties. \textbf{Clarity} evaluates the use of standard medical terminology together with syntactic completeness and discourse coherence. Pairs scoring below three on any axis are discarded, and a random portion of the remaining data is independently reviewed by two senior physicians to confirm the automatic ratings.

\subsection{Injecting Domain-Specific Knowledge}
\label{sec3.3}

\subsubsection{Data Collection}
\label{Data Collection}
Despite advancements in enhancing the model's capability to address clinical medical problems, the complexity and diversity of medical tasks still prevent the model from achieving satisfactory performance in specialized domains. Theoretically, the data heterogeneity across different domains underscores the urgent need to incorporate domain-specific knowledge into the model. To lay the foundation for enhancing interleaved reasoning capabilities, we collected 23 publicly available datasets related to chest X-rays, including ChestXray14, CheXpert, MIMIC-CXR, PadChest, RSNA, COVIDX-CXR-3, CXR-LT, BRAX, NLM-TB, Candid-PTX, BIMCV-COVID19, MS-CXR-T, VinDr-CXR, VinDr-PCXR, SIIM, MS-CXR, MIMIC-III, MIMIC-CXR-VQA, Rad-Restruct, MIMIC-NLE, RadGraph, MIMIC-Diff-VQA, and CXR-AL14. Each dataset typically contains one or more X-ray images from various perspectives, corresponding radiology reports, and common disease labels or visual localization coordinates. The diversity of data types is highly aligned with our objective of embedding X-ray-related knowledge into the model. Specifically, through comprehensive analysis of these extensive datasets, we identified three key capabilities that the model should possess: Visual Understanding, Text Generation, and Spatiotemporal Alignment.

\subsubsection{Construction of the CX-Set}
\label{Construction of the CX-Set}
To enhance the model’s performance across the three primary capabilities, we propose an instruction-fine-tuning dataset named CX-Set, tailored for chest X-rays. Specifically, we categorized different task types under each capability to comprehensively cover the knowledge domains in which the model should be trained. (i) \textbf{Visual Understanding:} Disease identification is the most common task in chest X-ray diagnostics, typically requiring differential diagnosis of 14 distinct diseases. Building on existing examination data, we constructed both close-ended and open-ended disease diagnosis questions, focusing on the interpretation and understanding of images. To diversify question types, close-ended questions are further classified into binary differentiation, single-disease diagnosis, and co-morbidity diagnosis. (ii) \textbf{Text Generation:} Writing radiology reports based on X-ray images is a time-consuming and repetitive task for clinicians. We aim to equip the model with the ability to generate specific sections of reports based on instructions, thereby alleviating the burden on clinicians. Specific tasks include findings generation, impression generation and findings summary. (iii) \textbf{Spatiotemporal Alignment:} This encompasses additional core functionalities expected of an "X-ray expert", such as matching images with radiology reports, distinguishing images from different perspectives, assessing disease progression, and localizing diseases. Collectively, these tasks comprehensively address the capabilities required of an exceptional chest X-ray expert model.

As shown in Figure \ref{fig:method}(b), We systematically constructed the CX-Set dataset based on the aforementioned tasks. To mitigate the impact of excessive noise on model performance, we first implemented an automated pipeline to clean the collected public datasets. This process involved removing data entries lacking images or radiology reports and filtering out cases with low-resolution images or reports that inadequately described disease conditions. Subsequently, to enhance data diversity, we designed 50 distinct question templates corresponding to the 13 previously defined task types. For task types where gold-standard answers could be directly extracted from the datasets, we adopted a "template-filling" approach for data construction \citep{chen2024vision}. Conversely, for tasks where answers could not be directly derived from existing data, we utilized DeepSeek-V3 \citep{liu2024deepseek} to generate standardized answers by summarizing radiology reports in alignment with specific questions. This automated data construction pipeline significantly reduced the labor-intensive manual annotation process. Leveraging this pipeline, we developed a large-scale chest X-ray instruction-fine-tuning dataset comprising 2,619,148 entries, covering 708,473 X-ray images. Each entry consists of three components: a task-specific question, one or more images, and the corresponding answer. To support enhanced interactive reasoning, we partitioned the entire dataset $D$ into two subsets: $D_{A}$ and $D_{R}$, which are used for injecting foundational X-ray knowledge and training subsequent reasoning capabilities, respectively. The objective of "knowledge injection" is to maximize the likelihood of generating correct answers based on the provided images and questions:
\begin{equation}
    \mathcal{L}_{\text{SFT}} = -\mathbb{E}_{(X, Q, A) \sim                 \mathcal{D}_A} \;\;\sum_{n=1}^{N} \;\; \log \pi_\theta(a_n \mid X, Q, a_{<n}),
    \tag{4}
\end{equation}
Where $X$, $Q$, and $A$ represent the image, question, and corresponding answer, respectively. $\pi_\theta$ denotes the policy model under training, $n$ represents the index of the token $a$ decoded by the model, and $N$ corresponds to the length of $A$.

\subsection{Cultivating Interleaved Reasoning Pattern}
\label{sec3.4}

\subsubsection{Definition of Interleaved Reasoning}
\label{Definition of Interleaved Reasoning}
Interleaved reasoning in medical MLLMs entails addressing complex, multi-step medical queries by resolving sub-tasks step-by-step. This approach decomposes a query into a series of intermediate reasoning steps, each producing a distinct, user-facing "sub-answer" that represents a confident, self-contained conclusion or milestone in the reasoning process. For example, in the diagnosis of a chest X-ray, a sub-answer might identify a specific abnormality, such as a lung nodule, as a resolved step, guiding subsequent reasoning toward differential diagnoses or treatment recommendations.

A critical distinction in this framework is between thinking and answering. Thinking refers to the internal, private reasoning processes—such as hypothesis generation or cross-modal data integration that are not directly accessible or immediately useful to the user \citep{xie2025interleaved}. In contrast, answering involves generating public, finalized outputs that advance the user’s understanding or support clinical decision-making. These outputs, presented as sub-answers, are conclusive at their respective stages and collectively contribute to a comprehensive response. For instance, in a multi-hop medical query requiring both image analysis and clinical correlation, a sub-answer might confirm a visual finding (e.g., "The X-ray reveals pulmonary edema") before proceeding to evidence-based differential diagnoses supported by image analysis. By interleaving reasoning across modalities and delivering clear, incremental sub-answers, medical MLLMs enhance transparency, interpretability, and collaboration with clinicians.

Specifically, we designed distinct interleaved reasoning modes based on the type of question in Figure \ref{method}(c). For close-ended questions, \textbf{CX-Mind} systematically evaluates each option to ensure that it is either retained or excluded based on evidence. Subsequently, it summarizes the reasoning process and intermediate results to derive the final conclusion. Conversely, when addressing open-ended questions, \textbf{CX-Mind} first identifies potential diseases based on a preliminary analysis of the image. It then evaluates each of these diseases through an evidence-based approach, ultimately providing a diagnostic conclusion. Formally, the interleaved reasoning mode is defined as follows:
\begin{equation}
   \mathcal{Y}_{close} = \underbrace{\tau^{\langle1\rangle} \oplus \rho^{\langle1\rangle}}_{\text{Option A}} \oplus \underbrace{\tau^{\langle2\rangle} \oplus \rho^{\langle2\rangle}}_{\text{Option B}} \oplus \dots \oplus \underbrace{\tau^{\langle s\rangle} \oplus \rho^{\langle s\rangle}}_{\text{Final Answer}}
\end{equation}

\begin{equation}
   \mathcal{Y}_{open} = \underbrace{\tau^{\langle1\rangle} \oplus \rho^{\langle1\rangle}}_{\text{Possible List}} \oplus \underbrace{\tau^{\langle2\rangle} \oplus \rho^{\langle2\rangle}}_{\text{Disease 1}} \oplus \dots \oplus \underbrace{\tau^{\langle s\rangle} \oplus \rho^{\langle s\rangle}}_{\text{Final Answer}}
\end{equation}
where $\mathcal{Y}$ represents the interleaved reasoning process, and $\tau$ and $\rho$ denote the reasoning and answer segments, respectively, corresponding to Equation \ref{equation2}.

\subsubsection{Cold Start for Interleaved Pattern}
\label{Cold Start for Interleaved Pattern}

To enhance the interleaved reasoning capabilities of \textbf{CX-Mind}, we initially employed a cold-start phase to enable the model to familiarize itself with and master the interleaved reasoning mode. This approach lays the groundwork for subsequent RL to further elevate reasoning performance. Specifically, we constructed 42,828 high-quality interleaved CoT data entries, guided by tailored prompts and supplemented by authentic radiology reports as supervisory signals. These data encompass complex tasks, including various types of disease identification and disease progression. The CoT data construction pipeline is outlined as follows:

\textbf{(i) Report Screening:} As authentic radiological findings in reports provide critical support for addressing complex tasks, we extracted information from the "Findings" section. However, the absence of a standardized format in the collected clinical reports posed challenges to extracting specific sections. To streamline this process, we targeted only reports containing the strings "FINDINGS:" and "IMPRESSION:", extracting the content between these markers to serve as supervisory signals for subsequent steps.

\textbf{(ii) Token Statistics:} More detailed radiological findings offer richer supportive evidence to guide the generation of high-quality CoT data. Accordingly, we utilized the tokenizer from the Qwen3 series to perform token counting on the radiology reports. Data entries with a "Findings" section exceeding 120 tokens were selected as the raw data for CoT construction.

\textbf{(iii) Category Balancing:} Statistical analysis revealed an imbalanced label distribution in the collected data. For instance, co-morbidity diagnosis tasks included a disproportionate number of entries labeled "No Finding." To enhance the robustness of reasoning capabilities, we employed an automated approach to balance the label distribution across different task types.

\begin{figure}[ht]
    \centering 
    \includegraphics[width=0.9\textwidth]{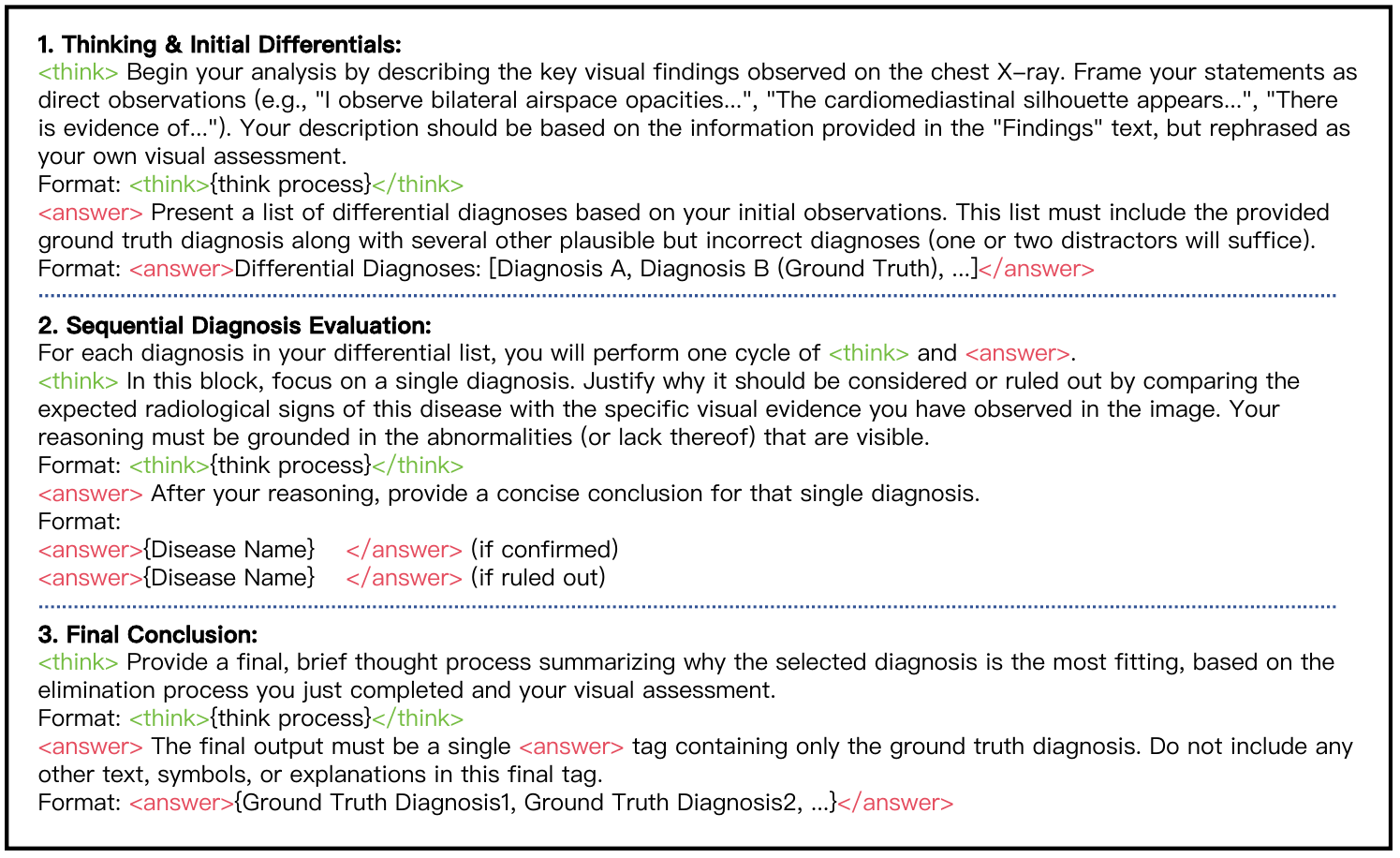} 
    \caption{Illustration of the interleaved reasoning process required for open-ended questions.} 
    \label{prompt1} 
\end{figure}

\textbf{(iv) Interleaved Reasoning Generation:} Leveraging the advanced language and vision capabilities of ChatGPT-4o, we generated interleaved reasoning processes for various task types. Notably, unlike other CoT construction methods, we utilized the authentic radiological findings extracted earlier as supervisory signals to minimize hallucinations. Specifically, we applied distinct interleaved reasoning logics tailored to different task types to align with practical scenarios. The reasoning logic for open-ended questions is illustrated in Figure \ref{prompt1}:

\begin{itemize}
    \item - Open-Ended: Thinking \& Initial Differentials \(\longrightarrow\) Sequential Diagnosis Evaluation \(\longrightarrow\) Final Conclusion
    \item - Close-Ended: Sequential Option Evaluation \(\longrightarrow\) Final Conclusion
    \item - Binary: Analysis and Conclusion
\end{itemize}

During the cold-start phase, we employed SFT to enable \textbf{CX-Mind} to adapt to the interleaved reasoning mode by utilizing a hybrid dataset composed of ${D}_{A}$ and a subset of ${D}_{R}$. The dataset $D_{R}$ was randomly divided into $D_{R1}$ and $D_{R2}$, each comprising half of the data. The former was utilized for cold-start training, with the objective function defined as follows:
\begin{equation}
    \mathcal{L}_{\text{Cold-Start}} = -\mathbb{E}_{(X, Q, C, R) \sim                 \mathcal{D}_{R1}} \;\;\sum_{n=1}^{N} \;\; \log \pi_\theta(r_n \mid X, Q, C, r_{<n}),
    \tag{7}
\end{equation}
where $C$ and $R$ represent the instruction output in a specific format and the complete interactive reasoning process, respectively.

\subsection{Curriculum-based Reinforcement Learning with Process Supervision}
\label{sec3.5}

This section details our curriculum‑based reinforcement learning pipeline. We adopt a GRPO backbone and extend it with step‑wise process rewards so that the policy is updated not only on the final answer but after every think–answer pair. Rewards are shaped to capture local consistency, cross‑turn coherence and medical safety within the interleaved stream. Training proceeds in two phases: the policy is first stabilised on close-ended questions and then unlocked for open‑ended queries, enabling a smooth transition from constrained to free‑form reasoning.

\subsubsection{GRPO‑Based Optimization Methodology}

With the release of Deepseek‑R1, GRPO has been shown to markedly enhance the reasoning capabilities of LLMs. Recent studies further demonstrate that the GRPO algorithm can improve the cross‑modal reasoning ability of medical vision–language models. As a variant of Proximal Policy Optimization (PPO), GRPO computes the generalized advantage estimate using group‑relative rewards. Unlike methods that rely on pretrained reward models, GRPO evaluates outputs through a verifiable rule‑based procedure. Specifically, for each sample $\bigl(
        \mathbf{X}_i,\,
        \mathbf{C}_i,\,
        \mathbf{Q}_i,\,
        \mathbf{R}_i,\,
        \mathbf{S}_i,\,
        \mathbf{A}_i
      \bigr)$ in $D$,
the model generates $G$ candidate responses $\{o_i\}_{i=1}^{G}$ under the current policy $\pi_\theta$. Each response receives a corresponding reward $r_i$, and the group of rewards is normalized to compute the advantage:
\begin{equation}
    A_i = \frac{r_i - \operatorname{mean}\!\bigl(\{r_j\}_{j=1}^G\bigr)}{\operatorname{std}\!\bigl(\{r_j\}_{j=1}^G\bigr)}.
\end{equation}
The GRPO objective can be written as:
\begin{equation}
J(\theta) = \mathbb{E}_{q\sim P(Q),\{o_i\}}\Biggl[
\frac{1}{G}\sum_{i=1}^G \min\bigl(R_i A_i, \;\mathrm{clip}(R_i, 1-\epsilon, 1+\epsilon)A_i\bigr)
\; - \beta\, D_{\mathrm{KL}}\bigl(\pi_{\theta} \| \pi_{\mathrm{ref}}\bigr)
\Biggr],
\end{equation}
where
$
R_i = \frac{\pi_{\theta}(o_i\mid \mathbf{X},\mathbf{C},\mathbf{Q})}{\pi_{\theta_{\mathrm{old}}}(o_i\mid \mathbf{X},\mathbf{C},\mathbf{Q})},
$
and its corresponding advantage $A_i$. The clipping mechanism limits extreme updates by constraining $R_i$ to the interval $[1-\epsilon,1+\epsilon]$, while the KL‑divergence term weighted by $\beta$ penalizes large deviations from the reference policy $\pi_{\mathrm{ref}}$.

\subsubsection{Rule-based Rewards Policy for Interleaved Process}
\begin{figure}[ht]
    \centering 
    \includegraphics[width=0.9\textwidth]{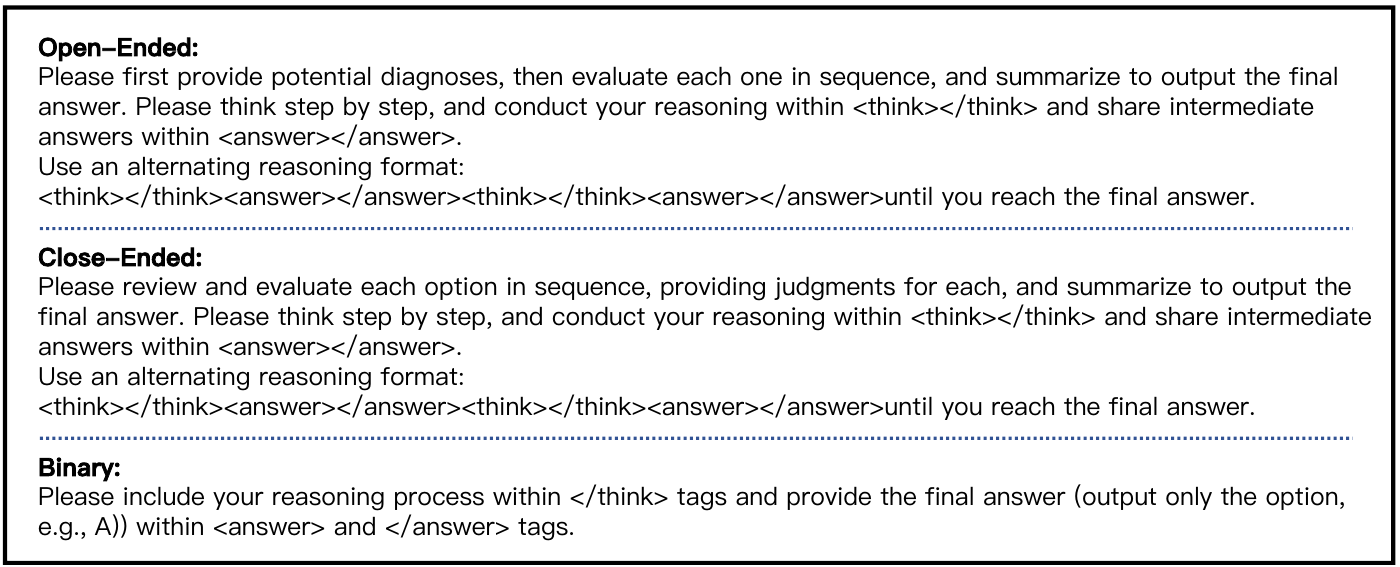} 
    \caption{Prompts following specific thinking-answer formats for different types of questions.} 
    \label{prompt2} 
\end{figure}

Because GRPO lacks inherent supervision on intermediate steps and pre‑trained reward models are prone to reward hacking, we define explicit rule‑based signals for both close‑ended and open‑ended questions in Figure \ref{fig:method}(d). To ensure correct parsing of the interleaved response, we prompt the model during training with the instruction template shown in Figure \ref{prompt2}.

\begin{itemize}
    \item \textbf{Format Reward} $R_{\mathrm{format}}$: A trajectory earns 1.0 if every <think>–</think> fragment is immediately followed by a matching <answer>–</answer> fragment and the sequence ends with an <answer>, otherwise 0.0.
    \item \textbf{Final‑result Reward} $R_{\mathrm{final}}$: The final accuracy reward evaluates the correctness of the last <answer> segment. For close‑ended questions we use a binary reward:
\begin{equation}
 R_{\text{final}}^{(\mathrm{closed})} =
    \begin{cases}
    1, & \text{if } \hat{y}_{\text{opt}} = y_{\text{gold}},\\[4pt]
    0, & \text{otherwise},
    \end{cases}
\end{equation}

where $\hat{y}_{\text{opt}}$ is the predicted choice and $y_{\text{gold}}$ is the ground‑truth label for a multiple‑choice question. For open‑ended diagnosis, a single choice is often insufficient because multiple conditions may
co‑occur. We therefore employ the micro‑F1 score, which balances precision and recall over the
entire disease set:
\begin{equation}
R_{\text{final}}^{(\mathrm{open})} =
\operatorname{F1}\bigl(\hat{y}_{\text{disease}},\, y_{\text{gold}}\bigr), 
\end{equation}

where $\hat{y}_{\text{disease}}$ and $y_{\text{gold}}$ denote the predicted and reference
disease sets, respectively.

\item \textbf{Process Reward $R_\mathrm{proc}$}: To harness real reports for supervising multi‑step reasoning and curb hallucinations, we first verify the final answer and only then allocate rewards to intermediate steps. Blindly rewarding the whole chain often fails because of the credit‑assignment problem. Instead, our mechanism first verifies the answer and then assigns an additional bonus to logically consistent and well‑structured reasoning traces. Inspired by \citep{xie2025interleaved}, we introduce a conditional, interleaved reward scheme that encourages correct reasoning. The method is entirely rule‑based and multi‑step, requiring no trained reward model, which makes it both verifiable and efficient. 

Concretely, a process reward is issued for close-ended questions only when three priors hold: (i) the output format is correct, (ii) the final answer is correct, and (iii) the mini‑batch accuracy exceeds the exponential moving average (EMA) of all previous batches.
This design ensures that intermediate supervision is applied only after the model shows meaningful learning progress. For open-ended questions, conditions (ii) and (iii) are combined by replacing accuracy with an F1‑score. Formally, let $\mathrm{Acc}_{b}$ be the accuracy of mini‑batch $b$ and $\mathrm{EMA}_{b-1}$ the EMA from preceding batches. Only samples satisfying all conditions receive the additional rewards $r_{\text{think}}$ and $r_{\text{ans}}$ for their intermediate reasoning steps and corresponding answers. The conditional process reward is:

\begin{equation}
\begin{split}
R_{\text{proc}}\!\bigl(\{\hat\tau_{i}^{\langle k\rangle},\hat\rho_{i}^{\langle k\rangle}\}_{k=1}^{s-1},\{\tau_{i}^{\langle k\rangle},\rho_{i}^{\langle k\rangle}\}_{k=1}^{s-1},\hat{y},y\bigr) \\
=\, \mathbb{I}(C)\,
    \sum_{k=1}^{s-1}
        r_\mathrm{think}\bigl(\hat\tau_{i}^{\langle k\rangle},\tau_{i}^{\langle k\rangle}\bigr) 
      +\, r_\mathrm{ans}\{\hat\rho_{i}^{\langle k\rangle},\rho_{i}^{\langle k\rangle}\}_{k=1}^{s-1},
\end{split}
\end{equation}

\begin{equation}
C =
  \mathrm{Check}\bigl(\{\hat\tau_{i}^{\langle k\rangle},\hat\rho_{i}^{\langle k\rangle}\}_{k=1}^{s}\bigr)
  \;\land\;
  \mathrm{Judge}(\hat{y})
  \;\land\;
  \bigl(\mathrm{Acc}_{b}>\mathrm{EMA}_{b-1}\bigr),
\end{equation}
where $\mathbb{I}(\cdot)$ denotes the indicator function, $\hat{y}$ is the predicted‑answer segment, and $\mathrm{Check}(\cdot)$ and $\mathrm{Judge}(\cdot)$ verify the interleaved format and the correctness of the final answer, respectively. For close-ended questions, $\mathrm{EMA}_{b-1}$ is iteratively updated with each batch’s accuracy. For open-ended questions, the same criterion applies with accuracy replaced by the batch‑level F1‑score.

For the thinking reward $r_{\mathrm{think}}$, the ground‑truth reasoning text is drawn verbatim from authoritative medical reports. We therefore devise a text-similarity-based reward function: each model‑generated reasoning part is aligned with its ground‑truth counterpart, after which BLEU‑1 and ROUGE‑L are computed to quantify lexical overlap. Their weighted average constitutes the reward:
\begin{equation}
    r_\mathrm{think}
    =\alpha\cdot\mathrm{BLEU}_1\bigl(\hat\tau_{i}^{\langle k\rangle},\tau_{i}^{\langle k\rangle}\bigr)
    +(1-\alpha)\cdot\mathrm{ROUGE}_L\bigl(\hat\tau_{i}^{\langle k\rangle},\tau_{i}^{\langle k\rangle}\bigr),
\end{equation}
where $\alpha\in[0,1]$ balances the two metrics.

For $r_{\mathrm{ans}}$, we adopt an All‑or‑None self‑consistent reward: a fixed coefficient $\gamma$ is granted only when every intermediate conclusion fragment is semantically identical to the reference answer, and no extra score is given if any discrepancy exists. 
\begin{equation}
r_\mathrm{ans}\bigl(\hat\rho_{i}^{\langle k\rangle},\rho_{i}^{\langle k\rangle}\bigr) = 
    \begin{cases}
    0.2, & \text{if } \hat\rho_{i}^{\langle k\rangle}=\rho_{i}^{\langle k\rangle},\;\forall\;k\in [1,s-1],\\[4pt]
    0, & \text{otherwise}.
    \end{cases}
\end{equation}

This minimalist rule constrains answer-level accuracy while preventing misleading incentives arising from partial matches.

    \item \textbf{Total Reward $R$}: The overall per‑trajectory reward can be formulated as the weighted sum:
    \begin{equation}
        R=\lambda R_{\mathrm{fomat}}+(1-\lambda)R_{\mathrm{final}}+R_{\mathrm{proc}},
    \end{equation}

Specifically, $\lambda$ modulates the relative contributions of the format reward and the final reward, where $R_{\mathrm{final}}$ is instantiated as $R_{\text{final}}^{(\mathrm{closed})}$ for close‑ended questions and as $R_{\text{final}}^{(\mathrm{open})}$ for open‑ended questions.

\end{itemize}

\subsubsection{Training Strategy}

In contrast to fields such as mathematics or formal logic, where the answer space is strictly defined, the reward landscape of open‑ended medical diagnosis is extremely sparse, which greatly reduces the efficiency of RL updates. Drawing on curriculum learning \citep{bengio2009curriculum} and further inspired by the approach in \citep{rui2025improving}, we first train the agent on simpler close‑ended tasks, such as binary judgements and multiple‑choice questions, to establish fundamental reasoning skills. We then refine the policy on open‑ended diagnostic tasks. The complete RL pipeline is summarised in Algorithm~\ref{alg:grpo_curriculum}.

\begin{algorithm}[tbp]
  \caption{Curriculum GRPO Training with Rule‑Based Process Rewards}
  \label{alg:grpo_curriculum}
  \begin{algorithmic}[1]
    \Require \textbf{Datasets:} close‑ended $\mathcal{D}_{\mathrm{closed}}$ and open‑ended $\mathcal{D}_{\mathrm{open}}$;  
            \textbf{Policies:} initial $\pi_{\theta}$ and reference policy $\pi_{\mathrm{ref}}$;  
            \textbf{GRPO hyper‑parameters:} group size $G$, clipping window $\epsilon$, KL weight $\beta$;  
            \textbf{Reward weights:} format–final trade‑off $\lambda$, thinking‑similarity weight $\alpha$, process bonus $\gamma$;  
            \textbf{Training steps:} $N_{\text{c}}$ (closed phase) and $N_{\text{o}}$ (open phase).
    \Statex
    \textcolor{blue}{// \textit{Method}}
    \Function{TrainPhase}{$\mathcal{D},\,N,\,\textsc{Closed}$}
      \State Initialise $\mathrm{EMA}\leftarrow 0$
      \For{$t=1$ \textbf{to} $N$}                                    \Comment{Iterate Mini‑batches}
        \State Sample batch $B\subset\mathcal{D}$;  $\;L_{\text{PG}}\leftarrow 0$
        \ForAll{query $q\in B$}
          \State Generate $G$ trajectories $\{o_j\}_{j=1}^{G}\sim\pi_{\theta}(o\mid q)$
          \For{$j=1$ \textbf{to} $G$}                                \Comment{Rule‑based Reward}
            \State $R_{\mathrm{format}}\!\leftarrow\!
                   \mathbb{I}\bigl(\text{interleaved format correct}\bigr)$
            \If{\textsc{Closed}}                                     \Comment{Final-result Reward}
              \State $R_{\mathrm{final}}\!\leftarrow\!
                     \mathbb{I}\bigl(\hat y_{\text{opt}} = y_{\text{gold}}\bigr)$
            \Else
              \State $R_{\mathrm{final}}\!\leftarrow\!
                     \operatorname{F1}(\hat{y}_{\text{disease}},y_{\text{gold}})$
            \EndIf
            \State $C\!\leftarrow\!(R_{\mathrm{format}}=1)\,\land\,
                   (R_{\mathrm{final}}>0)\,\land\,
                   (\text{BatchMetric}>\mathrm{EMA})$
            \If{$C$}                                                \Comment{Process Reward}
              \State $R_{\mathrm{proc}}\!\leftarrow\!
                     \sum_{k=1}^{s-1}r_{\mathrm{think}}^{(k)}+
                         r_{\mathrm{ans}}
                       $
            \Else
              \State $R_{\mathrm{proc}}\leftarrow 0$
            \EndIf
            \State $r_j\leftarrow
                   \lambda\,R_{\mathrm{format}}
                   +(1-\lambda)\,R_{\mathrm{final}}
                   +R_{\mathrm{proc}}$
          \EndFor
          \State $\mu,\sigma\!\leftarrow\!\text{mean/std}(\{r_j\})$;\;
                 $A_j\!=\!(r_j-\mu)/\sigma$
          \State $L_{\text{PG}}\mathrel{+}= \frac{1}{G}\!
                 \sum_{j=1}^{G}
                 \min\!\bigl(
                   R_jA_j,\,
                   \mathrm{clip}(R_j,1-\epsilon,1+\epsilon)A_j
                 \bigr)$
        \EndFor
        \State $\theta\!\leftarrow\!
               \theta\!+\!\eta\nabla_{\theta}
               \bigl(L_{\text{PG}}/|B|
               -\beta D_{\mathrm{KL}}(\pi_{\theta}\|\pi_{\mathrm{ref}})\bigr)$
        \State Update $\mathrm{EMA}$ with current batch accuracy (or F1)
      \EndFor
    \EndFunction
    \Statex
    \textcolor{blue}{// \textit{Training Pipeline}}
    \State \textbf{Phase 1: close‑ended RL}
    \State \Call{TrainPhase}{$\mathcal{D}_{\mathrm{closed}},\,N_{\text{c}},\,\textsc{Closed}= \textbf{True}$}

    \State \textbf{Phase 2: Open‑ended RL}
    \State \Call{TrainPhase}{$\mathcal{D}_{\mathrm{open}},\,N_{\text{o}},\,\textsc{Closed}= \textbf{False}$}
  \end{algorithmic}
\end{algorithm}

\section{Experiments}

\subsection{Dataset}
To demonstrate the broad generalization capabilities of \textbf{CX-Mind}, we evaluated its multimodal chest X-ray abilities and foundational clinical language proficiency. Specifically, we conducted experiments across three designated capability categories, including visual understanding, text generation, and spatiotemporal alignment, utilizing tasks constructed from MIMIC-CXR \citep{johnson2019mimic} and CheXpert \citep{irvin2019chexpert} as training and in-domain evaluation datasets.
 For each task category, 2,000 samples were randomly selected as the test set, with the remainder used for training. Correspondingly, we established an out-of-domain test set using OpenI \citep{demner2016preparing}, following the same data construction methodology (500 samples per task category), to validate the model’s robust generalization. For the evaluation of language proficiency, we selected datasets \citep{liu2024medbench} from five categories to comprehensively assess \textbf{CX-Mind}, encompassing medical language understanding (CHIP-CDN, CMeEE), medical language generation (IMCS-V2-MRG), complex medical reasoning (DDx-basic), medical safety and ethics (MedSafety), and medical knowledge question-answering (MedHG, Med-Exam). These datasets were randomly split, with 90\% of the data allocated for training and the remaining 10\% for testing.

\subsection{Evaluation Metircs}
For the evaluation metrics, we employed distinct measures tailored to the task types. For close-ended questions, accuracy (Acc) was used as the primary metric. For open-ended questions, disease identification tasks were assessed using the Jaccard index and accuracy, with a threshold of Jaccard \(> 0.5\). For report generation tasks, we adopted standard natural language generation metrics, including BLEU, ROUGE-1, ROUGE-2, and ROUGE-L. Additionally, BERTScore was utilized to evaluate semantic similarity. For disease localization tasks, we used Intersection over Union (IoU) and accuracy, with a threshold of IoU \(> 0.5\).

\subsection{Implementation Details}
The entire training pipeline, encompassing SFT and RL, was conducted using eight 80GB H100 GPUs. During the SFT phase, we adopted QwenVL2.5-VL-7B-Instruct as the base model for \textbf{CX-Mind}, employing LoRA for fine-tuning, with AdamW as the optimizer, a learning rate of 1e-4 following a cosine decay schedule, a batch size of 16, and bfloat16 mixed precision. For the RL phase, we employed the EasyR1 framework to achieve efficient VLM training, leveraging the HybridEngine from veRL. The training utilized the model after SFT for full-parameter fine-tuning, with FlashAttention-2 implemented to enhance efficiency. The rollout batch size was set to 64, with a maximum input pixel limit of 4,194,304. The learning rate was configured at 1e-6, with a weight decay of 0.01 and a KL penalty coefficient $\beta$ of 0.01, with $\lambda=0.2$ and $\alpha=0.3$. For each sample, 10 completions were generated using vLLM with a sampling temperature of 1.0. RL training was conducted for 1 epoch on close-ended questions and 2 epochs on open-ended questions, totaling approximately 12 hours.

\subsection{Baselines}
To evaluate the performance of \textbf{CX-Mind}, we compare it against the following three categories of baseline models:
\textbf{(i) General MLLMs:} This category includes high-performance vision-language models designed for general-purpose tasks. We selected both closed-source models, such as GPT-4o \citep{hurst2024gpt} and GPT-4o-mini \citep{hurst2024gpt}, known for their robust performance across diverse benchmarks including visual question answering and reasoning, and open-source models, such as Qwen2.5-VL-7B-Instruct \citep{bai2025qwen2}, Qwen2.5-VL-72B-Instruct \citep{bai2025qwen2} and DeepSeek-VL2 \citep{wu2024deepseek}, which demonstrate advanced capabilities in image and video understanding, document parsing, and visual agent tasks.
\textbf{(ii) Medical MLLMs:} This category comprises vision-language models pretrained or fine-tuned on medical corpora, tailored for medical imaging and clinical tasks. We include HuatuoGPT-Vision-7B \citep{chen2024huatuogptvision}, which leverages medical-specific data for enhanced clinical understanding, CheXagent \citep{chen2024chexagent}, designed for chest X-ray analysis.
\textbf{(iii) Medical Reasoning MLLMs:} This category includes models specifically fine-tuned for advanced medical reasoning tasks, often incorporating techniques such as GRPO. We selected MedVLM-R1 \citep{pan2025medvlm}, optimized for medical reasoning, and ChestX-Reasoner \citep{fan2025chestx}, which focuses on reasoning tasks specific to chest X-ray diagnostics.
These baseline models provide a comprehensive comparison framework, encompassing general-purpose multimodal capabilities, medical-specific pretraining, and advanced medical reasoning, to rigorously evaluate the performance and generalization of \textbf{CX-Mind} across diverse chest X-ray tasks.

\subsection{Main Results}
\label{Main Results}

\subsubsection{Performance on Visual Understanding}
\label{Performance on Visual Understanding}

\begin{figure}[tbp]
    \centering 
    \includegraphics[width=0.95\textwidth]{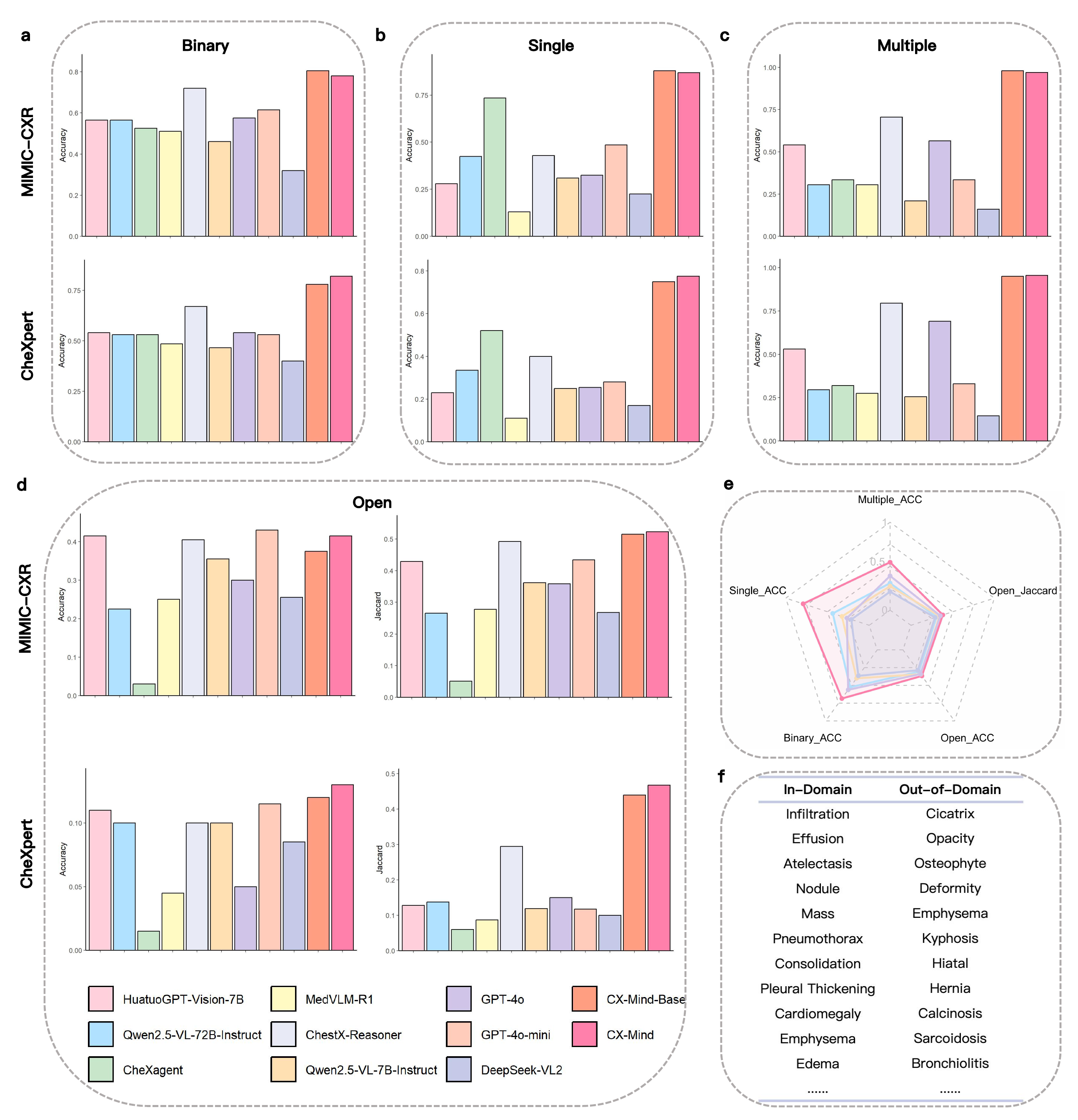} 
    \caption{Comparative study of CX-Mind and various baseline models in visual understanding capabilities. (a) Performance comparison of different models on a binary disease classification task, where the in-domain test set is primarily derived from authentic reports in MIMIC-CXR and CheXpert. (b) Performance comparison of different models on a single-disease identification task. (c) Performance comparison of different models on a co-morbidity identification task. (d) Performance comparison of different models on an open-ended disease identification task, evaluated using precision and Jaccard score. (e) Performance comparison of CX-Mind against mainstream models on an out-of-domain test set (OpenI). (f) Differences in disease categories covered by in-domain and out-of-domain test sets.} 
    \label{fig1} 
\end{figure}
We first evaluated the performance of \textbf{CX-Mind} in visual understanding capabilities, comparing it with other SOTA models. Specifically, close-ended and open-ended disease identification tasks were employed to assess the model’s ability to interpret X-ray images and detect abnormalities. As disease identification is a fundamental skill expected of radiologists, we categorized the tasks into binary (yes/no) classification, single-disease identification, co-morbidity identification, and free-form responses without predefined options, aiming to enhance the robustness of the evaluation results. Furthermore, we conducted evaluations on both in-domain (MIMIC-CXR, CheXpert) and out-of-domain (OpenI) test sets.

As shown in Figure \ref{fig1}, the results demonstrate that \textbf{CX-Mind} exhibits superior performance across various disease identification tasks, outperforming a range of large-scale closed-source and open-source models, such as GPT-4o and Qwen2.5-VL-72B-Instruct, as well as models fine-tuned on specific X-ray corpora, such as ChestX-Reasoner and CheXagent. This performance advantage remains consistent in out-of-domain test sets, underscoring \textbf{CX-Mind}’s robust visual understanding capabilities. Specifically, for binary classification tasks, \textbf{CX-Mind} outperformed the best baseline model, ChestX-Reasoner, by 15\% (82\% vs. 67\%) and 6\% (78\% vs. 72\%) in in-domain evaluations. In out-of-domain testing, \textbf{CX-Mind} achieved a 12.2\% performance improvement over GPT-4o. Due to the simplicity of binary tasks, \textbf{CX-Mind}’s advantage is less pronounced in this context. However, for single-disease and co-morbidity identification tasks, the performance gap widens significantly. Compared to models fine-tuned on chest X-ray datasets, \textbf{CX-Mind} achieved an average performance improvement of 19.5\% and 21\% over CheXagent and ChestX-Reasoner, respectively, in single-disease identification, and 63.5\% and 21.2\% in co-morbidity diagnosis. This highlights \textbf{CX-Mind}’s substantial advantage in complex tasks such as co-morbidity detection, an area often overlooked by comparable models.
For open-ended tasks, we required all models to output results in a list format, with ChatGPT-4o used to extract answers to account for non-standardized outputs. The results indicate that most models experience a significant performance drop when no options are provided, with the exception of \textbf{CX-Mind} and ChestX-Reasoner. Compared to the latter, \textbf{CX-Mind} achieved performance improvements of 3.1\% and 17.2\% on the MIMIC-CXR and CheXpert test sets, respectively.

\subsubsection{Performance on Text Generation}
\label{Performance on Text Generation}

\begin{figure}[ht]
    \centering 
    \includegraphics[width=1\textwidth]{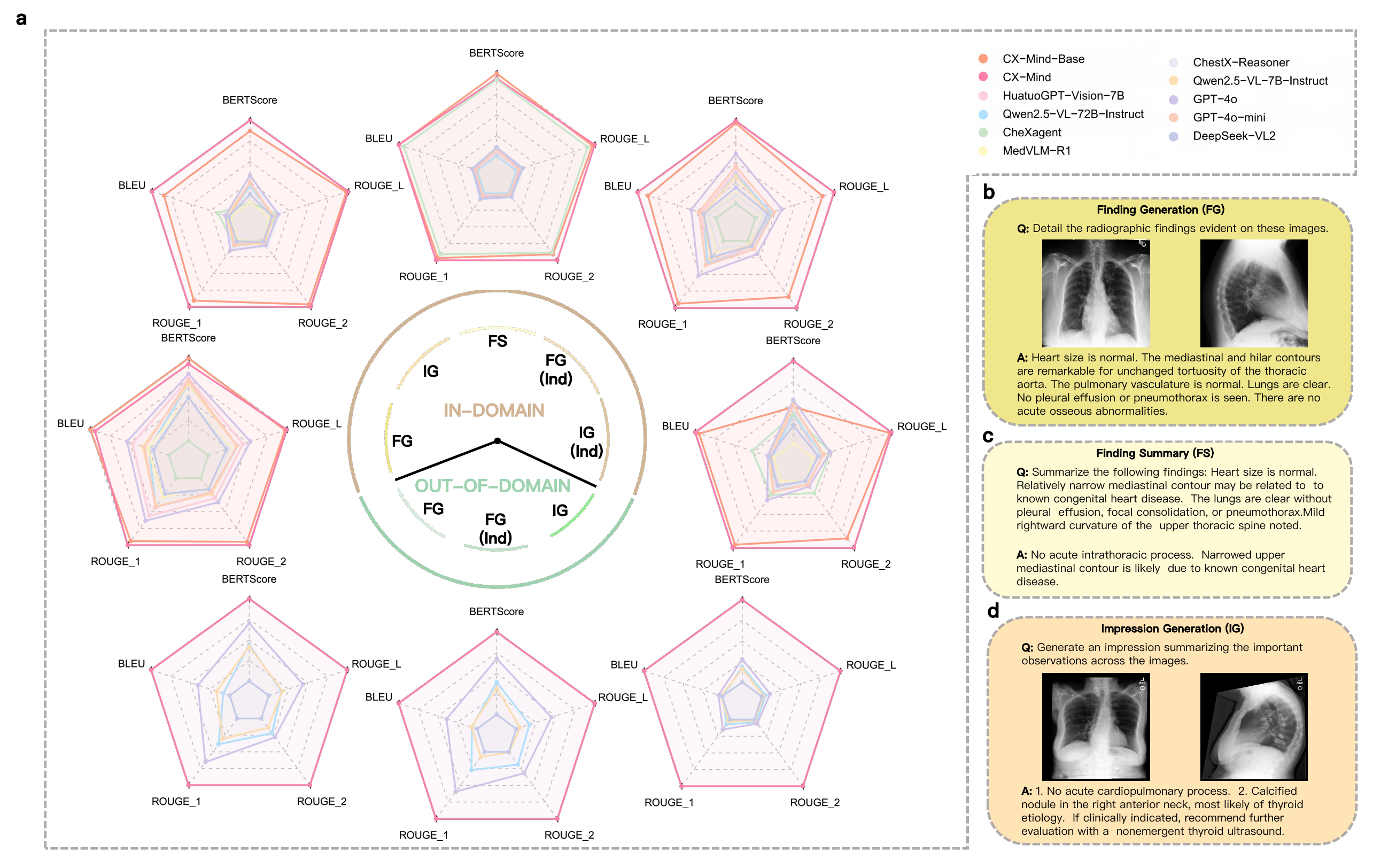} 
    \caption{Comparative study of CX-Mind and various baseline models in text generation capabilities. (a) We evaluated five in-domain and three out-of-domain tasks using five report generation metrics, where FG, IG, FS, FG(Ind), and IG(Ind) denote finding generation, impression generation, finding summarization, finding generation with indication, and impression generation with indication, respectively, representing five distinct text generation tasks. (b) Sample data demonstration for the finding generation task. (c) Sample data demonstration for the impression generation task. (d) Sample data demonstration for the finding summarization task.} 
    \label{fig2} 
\end{figure}
Writing radiology reports based on the interpretation of X-ray images is a labor-intensive and time-consuming repetitive task for clinicians. \textbf{CX-Mind}, through a four-stage progressive training approach, has achieved significant advancements in text generation capabilities. This progress contributes to alleviating clinicians' workload and enhancing the diagnostic efficiency of medical institutions.
To evaluate \textbf{CX-Mind}’s text generation performance, we conducted assessments across five tasks: finding generation, impression generation, and finding summarization, among others. To ensure a comprehensive evaluation, we not only focused on traditional metrics for report generation, such as character-level matching (e.g., BLEU, ROUGE), but also emphasized semantic similarity between the generated outputs and gold-standard reports.
Experimental results in Figure \ref{fig2}(a) demonstrate that \textbf{CX-Mind} achieves SOTA performance across various text generation tasks. Specifically, in the finding generation task, which requires robust anomaly detection capabilities, GPT-4o exhibited commendable performance compared to other baseline models. However, \textbf{CX-Mind} outperformed GPT-4o, achieving a 1.6\% higher BertScore, 7.6\% higher BLEU, and an average 11.1\% higher ROUGE score in the "Finding Generation" task. In the "Finding Generation (Indication)" task, \textbf{CX-Mind} surpassed GPT-4o by 3.6\% in BertScore, 21.7\% in BLEU, and an average of 22\% in ROUGE scores. These results underscore the model’s precise and reliable report generation capabilities.
For the finding summarization task, \textbf{CX-Mind} outperformed the best-performing baseline model, CheXagent, by an average of 2.2\% across all evaluation metrics. Additionally, the impression generation task, which requires high-level summarization of radiological findings from images, revealed a substantial performance gap between most models and \textbf{CX-Mind}. Notably, \textbf{CX-Mind}’s generated impressions demonstrated high semantic consistency with ground-truth reports, achieving BertScores of 90.3\% and 80.7\% in the "Impression Generation" and "Impression Generation (Indication)" tasks, respectively.
Furthermore, on an out-of-domain test set (OpenI), \textbf{CX-Mind} maintained SOTA performance in common report generation tasks, showcasing strong generalization capabilities.

\subsubsection{Performance on Spatiotemporal Alignment}
\label{Performance on Spatiotemporal Alignment}

\begin{figure}[ht]
    \centering 
    \includegraphics[width=1\textwidth]{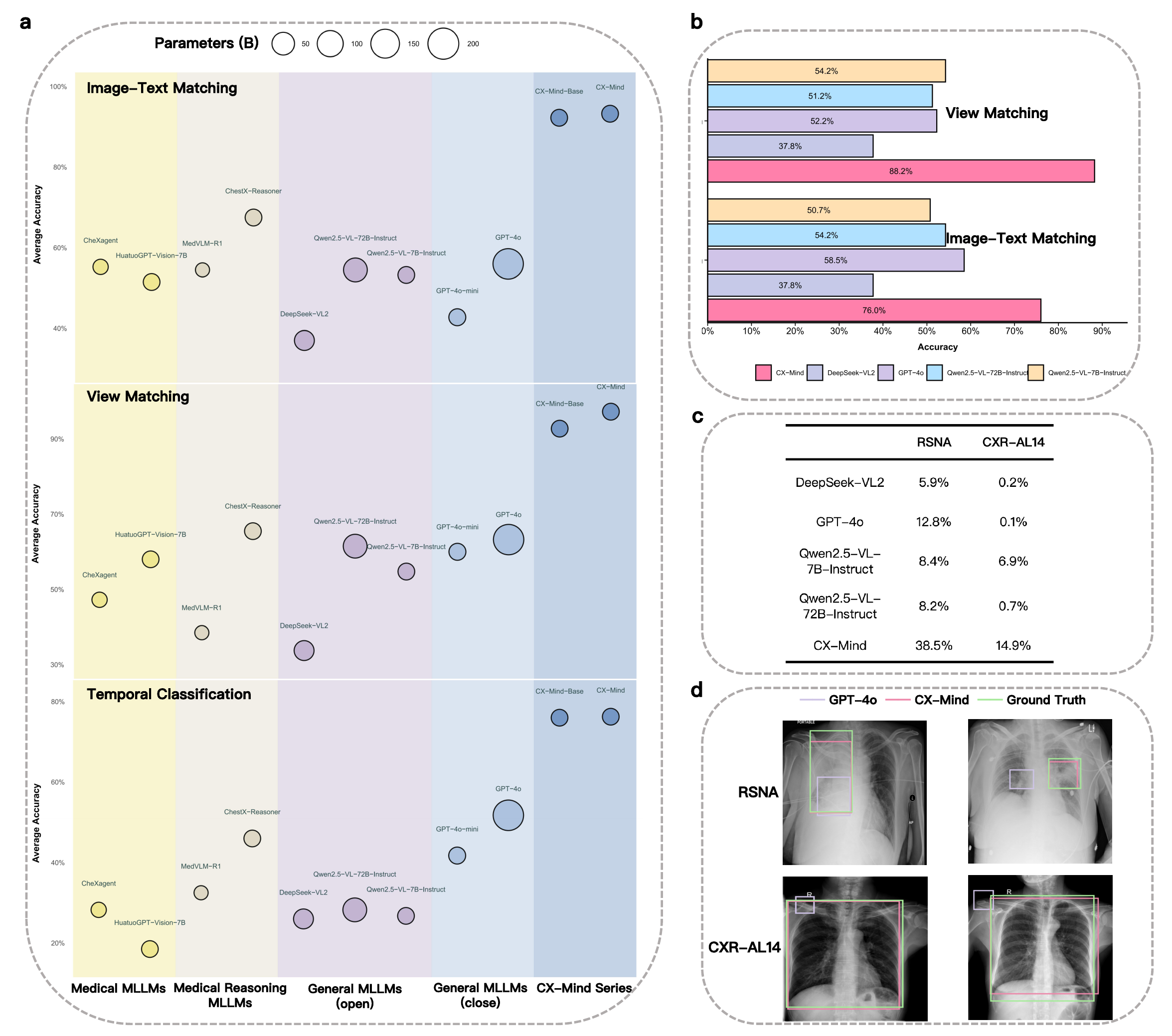} 
    \caption{Comparative study of CX-Mind and various baseline models in spatiotemporal alignment capabilities. (a) Comparison of average precision between CX-Mind and four other types of MLLMs across image-text matching, view matching, and temporal classification tasks, where circle size represents model parameter scale. (b) Performance comparison of CX-Mind against mainstream models on an out-of-domain test set. (c) Comparison of IoU for different models in the disease target detection task. (d) Visualization results of disease localization range for each model.} 
    \label{fig3} 
\end{figure}
In addition to the core capabilities outlined previously, an exceptional "X-ray expert" should possess robust temporal and spatial perception abilities. For temporal perception, we constructed two tasks—image-text matching and disease progression assessment—using the MIMIC-CXR and MS-CXR datasets. The image-text matching task involves selecting radiological examinations from the same patient at different time points to match the corresponding X-ray image. The disease progression task extracts two sequential radiology reports from the same patient, utilizing DeepSeek-V3 \citep{liu2024deepseek} to assess the progression of a specific critical disease, thereby evaluating the model’s temporal perception capabilities.
For spatial perception, we evaluated the model through image view classification and disease detection-localization tasks. Experimental results in Figure \ref{fig3}(a) reveal a substantial performance gap between baseline models and \textbf{CX-Mind} in temporal perception tasks. Our model outperformed the best-performing baseline, ChestX-Reasoner, by an average of 25.8\% in image-text matching and 30.2\% in disease progression tasks, demonstrating superior temporal perception capabilities.
In the simpler image view classification task, \textbf{CX-Mind} achieved near-perfect performance across multiple datasets. Compared to high-performing closed-source models GPT-4o and GPT-4o-mini, \textbf{CX-Mind} surpassed them by 31\% and 37.2\%, respectively. As shown in Figure \ref{fig3}(b), this significant advantage in spatiotemporal alignment was consistent in the out-of-domain OpenI test set, where \textbf{CX-Mind} achieved 76\% and 88.3\% performance in image-text matching and view classification tasks, respectively.
Furthermore, \textbf{CX-Mind} exhibited strong target localization capabilities, enabling the detection of potential anomalies in X-ray images. When compared to SOTA models with target detection functionality, \textbf{CX-Mind} achieved 38.5\% and 14.9\% in IoU metrics (Figure \ref{fig3}(d)) on the out-of-domain RSNA and CXR-AL14 datasets, respectively.

\subsubsection{Performance on Foundational Language Capabilities}
\label{Performance on Foundational Language Capabilities}

\begin{table}[t]
\caption{Comparative study of CX-Mind and various baseline models in foundational language capabilities. The upper half of the table includes LLMs with large parameter scales, while the lower half includes MLLMs based on the Qwen2.5-VL-7B-Instruct model.}
\vspace{6pt}
\label{tab:medical-benchmark}
\centering
\setlength{\tabcolsep}{2pt}
\renewcommand{\arraystretch}{1.2}
\begin{adjustbox}{width=\linewidth}
{\scriptsize
\begin{tabular}{
    lcccccccc
    }
\toprule
\textbf{Model} &
\multicolumn{2}{c}{\makecell{Medical\\Language\\Understanding}} &
\makecell{Medical\\Language\\Generation} &
\makecell{Complex\\Medical\\Reasoning} &
\makecell{Medical\\Safety \&\\Ethics} &
\multicolumn{2}{c}{\makecell{Medical QA}} &
\multirow{2}{*}{\makecell{Mean}} \\
\cmidrule(lr){2-3} \cmidrule(lr){7-8}
& CHIP-CDN & CMeEE & IMCS-V2-MRG & DDx-basic & MedSafety & MedHG & Med-Exam & \\
\midrule
o1-mini & 85.10 & 50.50 & 76.50 & 83.00 & 27.50 & 38.20 & 52.90 & 59.10 \\
\makecell[l]{Llama-3.3-70B-Instruct} & 98.60 & 48.80 & 91.00 & 84.20 & 66.90 & 71.10 & 66.80 & 75.34 \\
GPT-4o & 91.40 & 45.90 & 79.60 & 84.00 & 47.10 & 73.80 & 64.90 & 69.53 \\
\midrule
\makecell[l]{Qwen2.5-VL-7B-Instruct} & 88.67 & 39.86 & 86.56 & 75.89 & 18.00 & 40.21 & 16.33 & 52.22 \\
 CX-Mind (stage1) &
92.00\textsuperscript{\textcolor{red!50}{(+3.33)}} &
46.92\textsuperscript{\textcolor{red!70}{(+7.06)}} &
88.86\textsuperscript{\textcolor{red!50}{(+2.30)}} &
83.24\textsuperscript{\textcolor{red!70}{(+7.35)}} &
64.00\textsuperscript{\textcolor{red!100}{(+46.00)}} &
62.94\textsuperscript{\textcolor{red!90}{(+22.73)}} &
64.67\textsuperscript{\textcolor{red!100}{(+48.34)}} &
71.80\textsuperscript{\textcolor{red!90}{(+19.58)}} \\
 CX-Mind &
89.33\textsuperscript{\textcolor{red!30}{(+0.66)}} &
41.34\textsuperscript{\textcolor{red!40}{(+1.48)}} &
86.73\textsuperscript{\textcolor{red!20}{(+0.17)}} &
82.40\textsuperscript{\textcolor{red!70}{(+6.51)}} &
49.33\textsuperscript{\textcolor{red!95}{(+31.33)}} &
58.47\textsuperscript{\textcolor{red!85}{(+18.26)}} &
43.33\textsuperscript{\textcolor{red!90}{(+27.00)}} &
64.42\textsuperscript{\textcolor{red!80}{(+12.20)}} \\
\bottomrule
\end{tabular}
}
\end{adjustbox}
\end{table}

Fine-tuning MLLMs for visual tasks often leads to a degradation of their language capabilities. Consequently, we not only focused on \textbf{CX-Mind}’s performance in chest X-ray tasks but also prioritized maintaining its original language proficiency. This consistency is critical for addressing complex clinical medical tasks.
To this end, we collected seven publicly available text-only medical datasets, encompassing five key medical scenarios: language understanding, language generation, complex reasoning, safety and ethics, and knowledge-based question answering. In Table \ref{tab:medical-benchmark}, experimental results show that after the first stage of SFT on text-only medical data, \textbf{CX-Mind} (stage1) achieved performance comparable to closed-source models o1-mini and GPT-4o across most benchmarks. Notably, \textbf{CX-Mind} (stage1) outperformed these models by an average of 3.7\% on CHIP-CDN, 10.8\% on IMCS-V2-MRG, and 26.7\% on MedSafety. Compared to the SOTA text-only large language model Llama-3.3-70B-Instruct, which has ten times the parameter count, \textbf{CX-Mind} (stage1) exhibited only slightly inferior performance.
More importantly, \textbf{CX-Mind} maintained robust medical language capabilities even after X-ray knowledge injection and interleaved reasoning enhancement. Our findings indicate that \textbf{CX-Mind} outperformed the baseline model Qwen2.5-VL-7B-Instruct and the closed-source reasoning model o1-mini by an average of 12.2\% and 5.3\% across all benchmarks, respectively, without significant loss of precision compared to \textbf{CX-Mind} (stage1). These results collectively demonstrate that \textbf{CX-Mind} successfully enhances visual reasoning capabilities while preserving strong foundational language abilities.

\subsection{Ablation}

In this section, we first conduct a comprehensive ablation to disentangle the contribution of each training stage. Specifically, we isolate the four sequential modules and quantify their individual impact on downstream performance.  
Second, although SFT based on CoT data  has recently become the cold‑start recipe, the relative value of answer‑only samples versus CoT samples remains under‑explored for medical VLMs. We therefore compare both data regimes. Besides, we investigate how close‑ended and open‑ended samples shapes RL, revealing distinct gains that depend on task structure.  
Finally, we benchmark several RL variants that differ only in their reward definitions, confirming that a well‑designed reward signal is pivotal to stable policy improvement.  

\subsubsection{Training Stages}
\label{ablation:training}
In order to fully demonstrate the effectiveness of CuRL-VPR, we compare the performance of SFT stages 1, 2, 3 and RL stages on three types of benchmarks, and the results are shown in Table \ref{tab:ablation_stage}. To avoid the possible impact of extra reward policies, this part of the ablation experiments all use RL based on answer-only samples.

\paragraph{\textbf{The effectiveness and necessity of knowledge injection in SFT}}
As shown in Table \ref{tab:ablation_stage}, Stage 1 enhances language ability, the model’s text generation capability improves significantly without sacrificing visual understanding. For example, on the finding summary task, the average metric increases by 5.3\%. After Stage 2 large‐scale image fine‐tuning (Stage 1+2), the model’s overall performance increases from 27.3\% to 51.3\%. Further mixing CoT data and answer‐only data for SFT (Stage 1+2+3) pushes the model to continue improving across all three capabilities. Moreover, compared to the variant that applies RL without Stage 2 training (Stage 1+3+4), its overall performance drops significantly to 41.6\%, even below the variant without RL. This demonstrates that large‐scale image fine‐tuning not only establishes a robust visual‐semantic foundation but also provides a convergent initialization strategy for RL.

\paragraph{\textbf{Cold‐start RL further improves performance}}
RL is built upon a strong base model to comprehensively enhance the VLM’s interleaved reasoning capability. The Deepseek‐R0 style RL (Stage 1+2+4) achieves an overall score of 54.0\%, while with cold‐start, CX‑Mind‑Base combined with RL further increases by 2.7\%. This shows that cold‐start examples provide a better initialization for policy optimization, accelerating convergence and improving final performance.

\begin{center}
\setlength{\tabcolsep}{2pt}
\renewcommand{\arraystretch}{1}
\begin{scriptsize} 
\begin{longtable}{p{4.2cm}p{1.8cm}p{1.3cm}*{7}{C{1.2cm}}}
\caption{Ablation Results on training stages across tasks (higher is better). "Base" refers to Qwen2.5‑VL‑7B‑Instruct. "Stage" denotes the model after applying the corresponding training stage. CX‑Mind‑Base is the model trained through the full pipeline, where the RL stage uses only outcome (answer) rewards.
 "w/" denotes "with". "Ind." refers to Indication part in clinical report. "B1" and "B2" denote two benchmarks for the same task, distinguished by different question formulations. \textbf{Bold} highlights the best scores in each segment. }
 \\
\toprule
\textbf{Task} & \textbf{Dataset} & \textbf{Metric} &
\textbf{Base} & \textbf{Stage 1} & \textbf{Stage 1+2} &
\textbf{Stage 1+2+3} & \textbf{Stage 1+2+4} & \textbf{Stage 1+3+4} & \textbf{CX‑Mind -Base}\\
\midrule
\endfirsthead
\multicolumn{10}{c}{\scriptsize\slshape Continued from previous page}\\
\toprule
\textbf{Task} & \textbf{Dataset} & \textbf{Metric} &
Base & S1 & S1+2 & S1+2+3 & S1+2+4 & S1+3+4 & CX‑Mind\\
\midrule
\endhead
\midrule
\multicolumn{10}{r}{\scriptsize\slshape Continued on next page}\\
\midrule
\endfoot
\bottomrule
\endlastfoot
\multicolumn{10}{c}{\textit{Visual Understanding}}\\
\midrule
Disease Identification (open)   & CheXpert   & Accuracy & 10.0 & 11.5 & 11.5 & 11.5 & 11.5 & 11.0 & \textbf{12.0}\\
Disease Identification (open)   &    & Jaccard  & 11.9 & 11.5 & 12.5 & 21.8 & 43.0 & 20.7 & \textbf{43.9}\\
Disease Identification (open)   & MIMIC‑CXR  & Accuracy & 35.5 & 35.0 & 35.0 & 36.5 & 35.5 & 35.0 & \textbf{37.5}\\
Disease Identification (open)   &   & Jaccard  & 36.2 & 44.0 & 44.0 & 46.8 & 47.9 & 44.3 & \textbf{51.5}\\
Disease Identification (binary) & CheXpert   & Accuracy & 46.0 & 59.0 & \textbf{80.5} & 78.0 & 74.5 & 64.5 & \textbf{80.5}\\
Disease Identification (binary) & MIMIC‑CXR  & Accuracy & 46.5 & 61.0 & 72.0 & 77.0 & 77.5 & 65.0 & \textbf{78.0}\\
Disease Identification (single) & CheXpert   & Accuracy & 25.0 & 35.5 & 64.0 & 74.5 & 71.0 & 65.5 & \textbf{75.0}\\
Disease Identification (single) & MIMIC‑CXR  & Accuracy & 31.0 & 51.0 & 83.5 & 85.5 & 87.0 & 69.5 & \textbf{88.0}\\
Disease Identification (multiple) & CheXpert & Accuracy & 25.5 & 59.5 & 89.5 & 93.5 & 91.0 & 82.5 & \textbf{95.0}\\
Disease Identification (multiple) & MIMIC‑CXR & Accuracy & 21.0 & 52.5 & 84.0 & 95.0 & 90.5 & 62.5 & \textbf{98.0}\\
\midrule
\multicolumn{10}{c}{\textit{Text Generation}}\\
\midrule
Finding Generation                         & MIMIC‑CXR & BERTScore & 82.5 & 83.1 & 87.5 & 87.9 & 88.6 & 82.7 & \textbf{88.7}\\
Finding Generation                         &           & BLEU      &  7.2 &  8.2 & 16.7 & 17.0 & 17.3 & 12.0 & \textbf{19.7}\\
Finding Generation                         &           & ROUGE‑1   & 22.4 & 22.0 & 37.9 & 38.4 & 38.8 & 31.6 & \textbf{40.7}\\
Finding Generation                         &           & ROUGE‑2   &  4.2 &  4.9 & 15.6 & 15.9 & 16.0 &  9.8 & \textbf{16.1}\\
Finding Generation                         &           & ROUGE‑L   &  8.0 &  8.3 & \textbf{29.2} & 27.4 & 29.1 & 15.2 & 28.5\\
Finding Generation w/ Ind.            & MIMIC‑CXR          & BERTScore & 84.6 & 84.8 & 82.3 & 88.4 & 89.1 & 85.3 & \textbf{90.5}\\
Finding Generation w/ Ind.            &           & BLEU      &  7.0 &  7.8 & 29.9 & \textbf{31.6} & 30.6 & 10.4 & 30.4\\
Finding Generation w/ Ind.            &           & ROUGE‑1   & 21.4 & 22.9 & 47.2 & 48.5 & 48.5 & 23.0 & \textbf{49.7}\\
Finding Generation w/ Ind.            &           & ROUGE‑2   &  2.9 &  2.9 & 30.5 & 27.9 & \textbf{33.5} &  7.7 & 28.5\\
Finding Generation w/ Ind.            &           & ROUGE‑L   & 13.3 & 14.7 & 37.0 & 36.4 & \textbf{37.4} & 23.6 & \textbf{37.4}\\
Finding Summary                            &  MIMIC‑CXR         & BERTScore & 85.5 & 85.6 & 92.9 & 93.3 & 93.4 & 87.7 & \textbf{93.6}\\
Finding Summary                            &           & BLEU      &  7.5 &  8.4 & 45.3 & 46.1 & 46.8 & 30.1 & \textbf{47.4}\\
Finding Summary                            &           & ROUGE‑1   & 19.3 & 19.8 & 60.3 & 63.4 & 63.5 & 40.6 & \textbf{65.1}\\
Finding Summary                            &           & ROUGE‑2   &  9.1 &  9.9 & 48.6 & 50.9 & 51.9 & 30.7 & \textbf{53.1}\\
Finding Summary                            &           & ROUGE‑L   & 17.2 & 17.7 & 58.8 & 61.5 & 62.0 & 50.9 & \textbf{63.4}\\
Impression Generation                      & MIMIC‑CXR          & BERTScore & 83.6 & 83.4 & 88.0 & 88.4 & 89.1 & 84.1 & \textbf{89.2}\\
Impression Generation                      &           & BLEU      &  3.7 &  5.7 & 26.6 & 27.2 & 26.8 &  6.8 & \textbf{27.6}\\
Impression Generation                      &           & ROUGE‑1   & 10.0 & 10.7 & 36.8 & 38.7 & 38.1 & 14.1 & \textbf{44.5}\\
Impression Generation                      &           & ROUGE‑2   &  2.0 &  2.2 & 27.2 & 28.2 & 27.8 & 15.7 & \textbf{31.8}\\
Impression Generation                      &           & ROUGE‑L   &  7.2 &  5.5 & 36.3 & 37.8 & 37.5 & 10.6 & \textbf{42.8}\\
Impression Generation w/ Ind.         &  MIMIC‑CXR         & BERTScore & 48.4 & 48.6 & 55.1 & 55.0 & 55.4 & 50.1 & \textbf{78.3}\\
Impression Generation w/ Ind.         &           & BLEU      &  2.3 &  5.3 & 18.1 & 17.4 & 18.1 & 17.4 & \textbf{18.3}\\
Impression Generation w/ Ind.         &           & ROUGE‑1   &  7.0 &  7.8 & 20.5 & 19.1 & 21.0 & 13.9 & \textbf{21.8}\\
Impression Generation w/ Ind.         &           & ROUGE‑2   &  1.1 &  2.0 & 13.9 & 14.2 & \textbf{15.2} &  7.7 & \textbf{15.2}\\
Impression Generation w/ Ind.         &           & ROUGE‑L   &  5.9 &  7.1 & 20.3 & 19.1 & \textbf{21.9} & 10.5 & 21.5\\
\midrule
\multicolumn{10}{c}{\textit{Spatiotemporal Alignment}}\\
\midrule
Image–Text Match            & B1        & Accuracy & 47.5 & 48.0 & 81.0 & \textbf{92.0} & 86.0 & 75.0 & 90.5\\
Image–Text Match            & B2        & Accuracy & 59.0 & 60.0 & 92.5 & 92.5 & \textbf{94.5} & 81.5 & 94.0\\
Temporal Classification           & MS‑CXR    & Accuracy & 26.5 & 35.5 & 68.5 & 67.5 & 70.0 & 54.5 & \textbf{71.5}\\
Temporal Classification     & MIMIC‑CXR & Accuracy & 27.0 & 43.5 & \textbf{80.5} & 77.5 & 79.5 & 67.0 & \textbf{80.5}\\
View Match                  & B1        & Accuracy & 48.5 & 59.5 & 78.0 & \textbf{96.0} & 87.0 & 71.0 & 92.0\\
View Match                  & B2        & Accuracy & 61.0 & 63.0 & 63.0 & 69.0 & 69.0 & 64.0 & \textbf{93.5}\\
\midrule
\multicolumn{3}{l}{\textbf{Average}} &
27.3 & 31.9\textsuperscript{\textcolor{red!40}{+4.6}} &
51.3\textsuperscript{\textcolor{red!70}{+24.0}} &
53.5\textsuperscript{\textcolor{red!75}{+26.2}} &
54.0\textsuperscript{\textcolor{red!80}{+26.7}} &
41.6\textsuperscript{\textcolor{red!50}{+14.3}} &
\textbf{56.7}\textsuperscript{\textcolor{red!100}{+29.4}}\label{tab:ablation_stage}
\end{longtable}
\end{scriptsize}
\end{center}

\subsubsection{Training Data}

\begin{figure}[tbp]
    \centering 
    \includegraphics[width=1\textwidth]{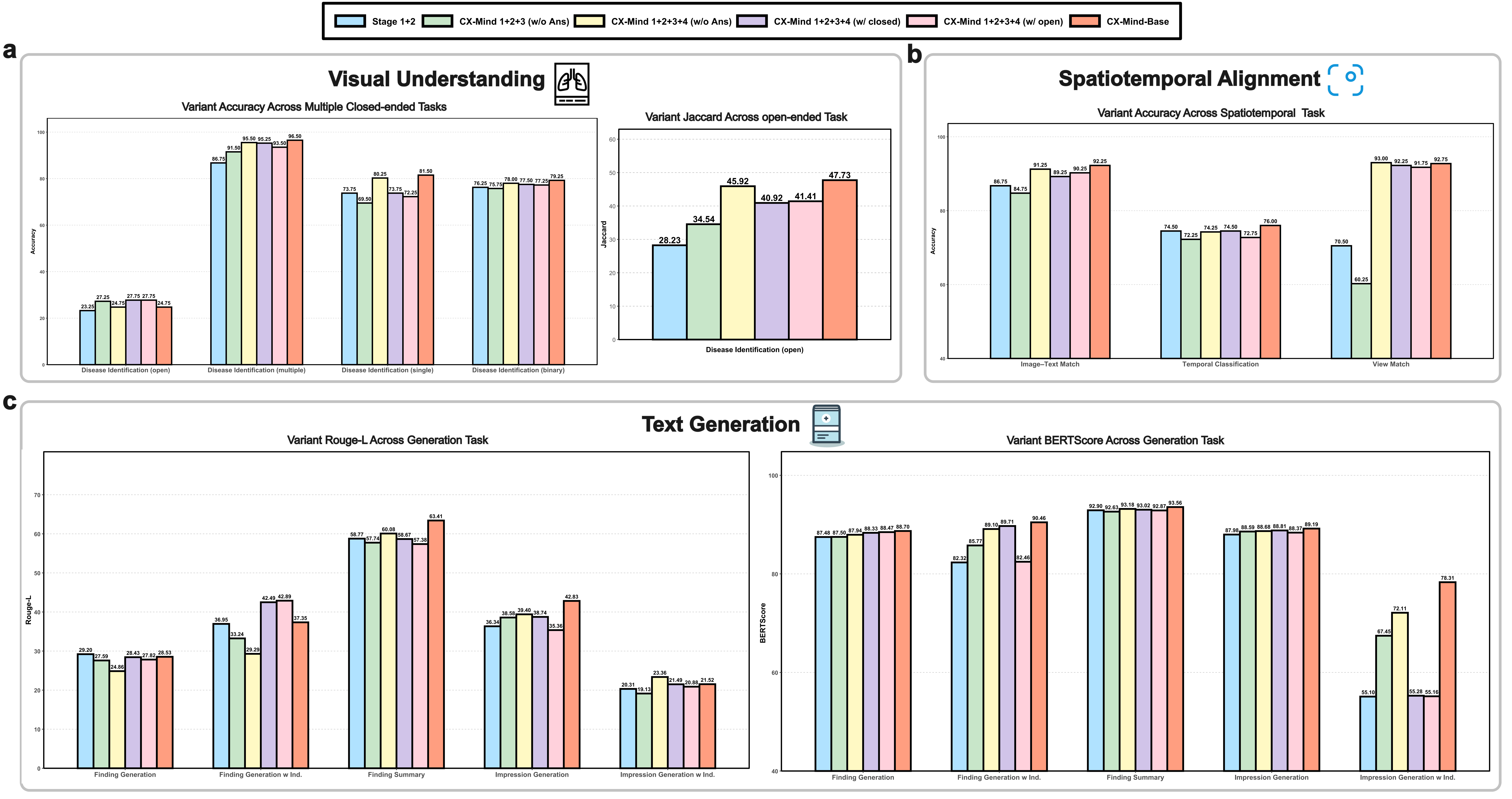} 
    \caption{
Ablation study on the impact of training data for our trained models, including \textbf{CX-Mind} stage1+2, \textbf{CX-Mind} 1+2+3 (w/o Ans), \textbf{CX-Mind} 1+2+3+4 (w/o Ans), \textbf{CX-Mind} 1+2+3+4 (w/ closed), \textbf{CX-Mind} 1+2+3+4 (w/ open), and \textbf{CX-Mind}-Base. 
\textbf{a}. Comparison of Visual Understanding capabilities among variants, covering five disease identification tasks evaluated using Accuracy and Jaccard metrics. 
\textbf{b}. Comparison of Visual Understanding capabilities among variants, including three localization and progression tasks.
\textbf{c}. Comparison of Text Generation capabilities among variants, covering five generation tasks.
}
    \label{fig:data ablation} 
\end{figure}

To demonstrate the effectiveness of each part of the training data, we performed ablations on both the cold‑start data and the curriculum‑based RL. Similarly, to avoid interference, all models undergoing RL here use only the final‑answer reward $R_\mathrm{final}$.

\paragraph{\textbf{Cold‑start with answer‑only supervision samples is important}}
To further validate the value of answer‑only samples in cold‑start and subsequent reinforcement learning, we also evaluated the performance when introducing answer‑only data in Stage 3 (variant CX‑Mind 1+2+3) and then applying RL on this basis (variant CX‑Mind 1+2+3+4 (w/ Ans)). As shown in Figure \ref{fig:data ablation}, using only CoT for cold‑start (CX‑Mind 1+2+3 (w/o Ans)) leads to performance degradation on certain tasks. For example, finding generation drops by 1.61\% compared to the Stage 2 baseline, and after adding RL (CX‑Mind 1+2+3+4 (w/o Ans)), the average accuracy further decreases by 0.857\% and the ROUGE‑L score by 3.33\% relative to CX‑Mind‑Base. In contrast, when we mix in answer‑only samples during Stage 3, the model’s capabilities are significantly restored and improved.

\paragraph{\textbf{Close‑ended and open‑ended RL further enhance reasoning capabilities}}
Compared to CX‑Mind-Base, omitting either close‑ended or open‑ended data during RL results in degraded performance. In visual understanding, disease identification Jaccard on open‐ended examples falls by 6.32\% without close‑ended rewards and by 6.81\% without open‐ended rewards. In text generation, the Rouge-L for finding summary decreases by 6.03\% when close‑ended rewards are removed and by 4.74\% when open‐ended rewards are omitted. Spatiotemporal alignment average accuracy drops by 2.1\%. Thus, both reward types are crucial for effective RL in interleaved cross‑modal reasoning.

\subsubsection{Different Reward Strategies}
We evaluate the effectiveness of different reward configurations in Table \ref{tab:ablation:reward}, using \textbf{CX-Mind}-Base trained solely on $D_{\mathrm{A}}$ as the baseline for comparison.

\paragraph{\textbf{Direct supervision at every step negatively impacts model performance}} 
Introducing direct step-wise thinking supervision (Direct Think) reduces the overall average metric from 56.7\% to 54.1\%, suggesting that unconstrained intermediate supervision may hinder effective model learning and exacerbate the credit assignment issue.

\paragraph{\textbf{Two additional process-based rewards are important}}
Incorporating conditional thinking rewards (Base (Think)) slightly raises the average performance to 56.9\% (+0.2\%). When further adding process-answer rewards during close-ended RL (w/ $r_{\text{ans}}^{(\text{open})}$) and open-ended RL stages (w/ $r_{\text{ans}}^{(\text{close})}$), the average metric improves further. Finally, \textbf{CX-Mind}, integrating all reward strategies and mechanisms, achieves the best overall performance across tasks. These results highlight the necessity of process-oriented rewards guided by real medical reports for multimodal interleaved diagnostic reasoning in medical domains.

\begin{center}
\begin{scriptsize}
\setlength{\tabcolsep}{2pt}
\renewcommand{\arraystretch}{1}
\begin{longtable}{p{4.3cm}p{1.9cm}p{1.4cm}*{6}{C{1.3cm}}}
\caption{Reward ablation across tasks (higher is better). "Direct Think" applies step-wise think rewards unconditionally at every training step. "Base (Think)" applies step-wise think rewards conditionally. In addition to (Think), "w/ \(\,r_{\mathrm{ans}}^{(\mathrm{open})}\,\)" and "w/ \(\,r_{\mathrm{ans}}^{(\mathrm{close})}\,\)" are variants that add step‑wise answer rewards during the open‑ended and close‑ended RL stages, respectively. "\textbf{CX‑Mind}" is the full model with both process and answer rewards. \textbf{Bold} highlights the best score in each row.}

\\
\toprule
\textbf{Task} & \textbf{Dataset} & \textbf{Metric} &
\textbf{CX‑Mind -Base} & \textbf{Direct Think} & \textbf{Base (Think)} &
\textbf{w/ $\boldsymbol{r_{\text{ans}}^{(\text{open})}}$} & \textbf{w/ $\boldsymbol{r_{\text{ans}}^{(\text{close})}}$} & \textbf{CX‑Mind}\\
\midrule
\endfirsthead
\multicolumn{9}{c}{\scriptsize\slshape Continued from previous page}\\
\toprule
\textbf{Task} & \textbf{Dataset} & \textbf{Metric} &
Base & Direct & Base(T) & –$R_{\text{close}}$ & –$R_{\text{open}}$ & CX‑Mind\\
\midrule
\endhead
\midrule
\multicolumn{9}{r}{\scriptsize\slshape Continued on next page}\\
\midrule
\endfoot
\bottomrule
\endlastfoot
\multicolumn{9}{c}{\textit{Visual Understanding}}\\
\midrule
Disease Identification (open)        & CheXpert   & Accuracy & 12.0 & 11.5 & 12.0 & 12.0 & 11.5 & \textbf{13.0}\\
Disease Identification (open)        &            & Jaccard  & 43.9 & 11.5 & 42.0 & 40.3 & 39.3 & \textbf{46.7}\\
Disease Identification (open)        & MIMIC‑CXR  & Accuracy & 37.5 & \textbf{44.0} & 40.0 & 41.0 & 40.5 & 41.5\\
Disease Identification (open)        &            & Jaccard  & 51.5 & 44.0 & 51.0 & 52.0 & 51.5 & \textbf{52.3}\\
Disease Identification (binary)      & CheXpert   & Accuracy & \textbf{80.5} & 77.5 & 77.0 & 77.5 & 77.5 & 78.0\\
Disease Identification (binary)      & MIMIC‑CXR  & Accuracy & 78.0 & 77.5 & 78.5 & 78.5 & 79.5 & \textbf{82.0}\\
Disease Identification (single)      & CheXpert   & Accuracy & 75.0 & 57.5 & 79.5 & \textbf{80.0} & 73.0 & 77.5\\
Disease Identification (single)      & MIMIC‑CXR  & Accuracy & 88.0 & 81.0 & 85.5 & \textbf{90.0} & 86.0 & 87.0\\
Disease Identification (multiple)    & CheXpert   & Accuracy & 95.0 & 94.5 & 94.5 & \textbf{96.0} & 94.5 & 95.5\\
Disease Identification (multiple)    & MIMIC‑CXR  & Accuracy & \textbf{98.0} & 96.0 & 96.0 & 96.5 & 97.0 & 97.0\\
\midrule
\multicolumn{9}{c}{\textit{Text Generation}}\\
\midrule
Finding Generation                    & MIMIC‑CXR  & BERTScore & \textbf{88.7} & 88.1 & 88.1 & 87.9 & 87.6 & 87.8\\
Finding Generation                    &            & BLEU      & \textbf{19.7} & 15.4 & 18.6 & 18.4 & 18.0 & 18.7\\
Finding Generation                    &            & ROUGE‑1   & \textbf{40.7} & 36.6 & 41.3 & 38.4 & 41.3 & 42.6\\
Finding Generation                    &            & ROUGE‑2   & \textbf{16.1} & 13.2 & 15.9 & 16.2 & 15.5 & 16.9\\
Finding Generation                    &            & ROUGE‑L   & \textbf{28.5} & 26.4 & 28.2 & 28.7 & 26.5 & 29.0\\
Finding Generation w/ Ind.            & MIMIC‑CXR  & BERTScore & 90.5 & 90.9 & 89.9 & 90.8 & 90.9 & \textbf{91.0}\\
Finding Generation w/ Ind.            &            & BLEU      & 30.4 & \textbf{32.8} & 33.9 & 33.4 & 33.3 & 34.2\\
Finding Generation w/ Ind.            &            & ROUGE‑1   & 49.7 & 51.4 & 48.9 & 51.2 & 52.1 & \textbf{52.0}\\
Finding Generation w/ Ind.            &            & ROUGE‑2   & 28.5 & 31.6 & 32.5 & 33.6 & \textbf{33.5} & 33.0\\
Finding Generation w/ Ind.            &            & ROUGE‑L   & 37.4 & 41.3 & 39.3 & 41.7 & \textbf{41.7} & 41.8\\
Finding Summary                       & MIMIC‑CXR  & BERTScore & \textbf{93.6} & 93.0 & 93.1 & 93.1 & 93.0 & 93.1\\
Finding Summary                       &            & BLEU      & \textbf{47.4} & 44.6 & 46.4 & 47.8 & 46.6 & 47.4\\
Finding Summary                       &            & ROUGE‑1   & \textbf{65.1} & 61.7 & 65.9 & 65.2 & 60.9 & 66.1\\
Finding Summary                       &            & ROUGE‑2   & \textbf{53.1} & 48.8 & 54.7 & 54.6 & 53.1 & 55.8\\
Finding Summary                       &            & ROUGE‑L   & \textbf{63.4} & 60.0 & 63.7 & 64.4 & 62.2 & 64.3\\
Impression Generation                 & MIMIC‑CXR  & BERTScore & \textbf{89.2} & 90.9 & 89.9 & 90.8 & 90.9 & 90.7\\
Impression Generation                 &            & BLEU      & \textbf{27.6} & 28.6 & 30.2 & 30.5 & 29.2 & 27.7\\
Impression Generation                 &            & ROUGE‑1   & 44.5 & 38.3 & 44.8 & 45.0 & 45.5 & \textbf{45.6}\\
Impression Generation                 &            & ROUGE‑2   & 31.8 & 27.8 & 31.2 & 28.7 & 30.3 & \textbf{32.7}\\
Impression Generation                 &            & ROUGE‑L   & 42.8 & 37.5 & 43.0 & 43.2 & 40.5 & \textbf{43.6}\\
Impression Generation w/ Ind.         & MIMIC‑CXR  & BERTScore & \textbf{78.3} & 59.8 & 71.2 & 76.2 & 78.8 & 80.7\\
Impression Generation w/ Ind.         &            & BLEU      & 18.3 & 32.9 & 19.0 & 14.1 & 21.3 & \textbf{22.7}\\
Impression Generation w/ Ind.         &            & ROUGE‑1   & 21.8 & 51.4 & 20.6 & 20.0 & 21.3 & \textbf{21.9}\\
Impression Generation w/ Ind.         &            & ROUGE‑2   & 15.2 & 32.7 & 15.9 & 14.4 & 16.4 & \textbf{17.4}\\
Impression Generation w/ Ind.         &            & ROUGE‑L   & 21.5 & 22.7 & 20.3 & 19.3 & 21.9 & \textbf{21.9}\\
\midrule
\multicolumn{9}{c}{\textit{Spatiotemporal Alignment}}\\
\midrule
Image–Text Match                      & B1        & Accuracy & 90.5 & 91.5 & 90.5 & 88.5 & 90.0 & \textbf{92.0}\\
Image–Text Match                      & B2        & Accuracy & 94.0 & 92.0 & 94.0 & 94.0 & 92.0 & \textbf{94.5}\\
Temporal Classification                       & MS‑CXR    & Accuracy & 71.5 & 71.5 & 71.0 & 71.0 & \textbf{72.0} & 72.0\\
Temporal Classification               & MIMIC‑CXR & Accuracy & 80.5 & 80.0 & \textbf{80.5} & 80.0 & 81.0 & 80.5\\
View Match                            & B1        & Accuracy & 92.0 & 92.0 & \textbf{95.0} & 95.0 & 95.5 & 95.5\\
View Match                            & B2        & Accuracy & 93.5 & 96.5 & 98.5 & 98.5 & \textbf{99.5} & 99.0\\
\midrule
\multicolumn{3}{l}{\textbf{Average}} &
56.7 & 54.1\textsuperscript{\textcolor{green!90}{-2.6}} &
56.9\textsuperscript{\textcolor{red!80}{+0.2}} &
57.1\textsuperscript{\textcolor{red!90}{+0.4}} &
56.8\textsuperscript{\textcolor{red!70}{+0.1}} &
\textbf{58.2}\textsuperscript{\textcolor{red!100}{+1.5}}\label{tab:ablation:reward}\\
\end{longtable}
\end{scriptsize}
\end{center}

\subsection{Clinical Evaluation}
\begin{figure}[tbp]
    \centering 
    \includegraphics[width=0.9\textwidth]{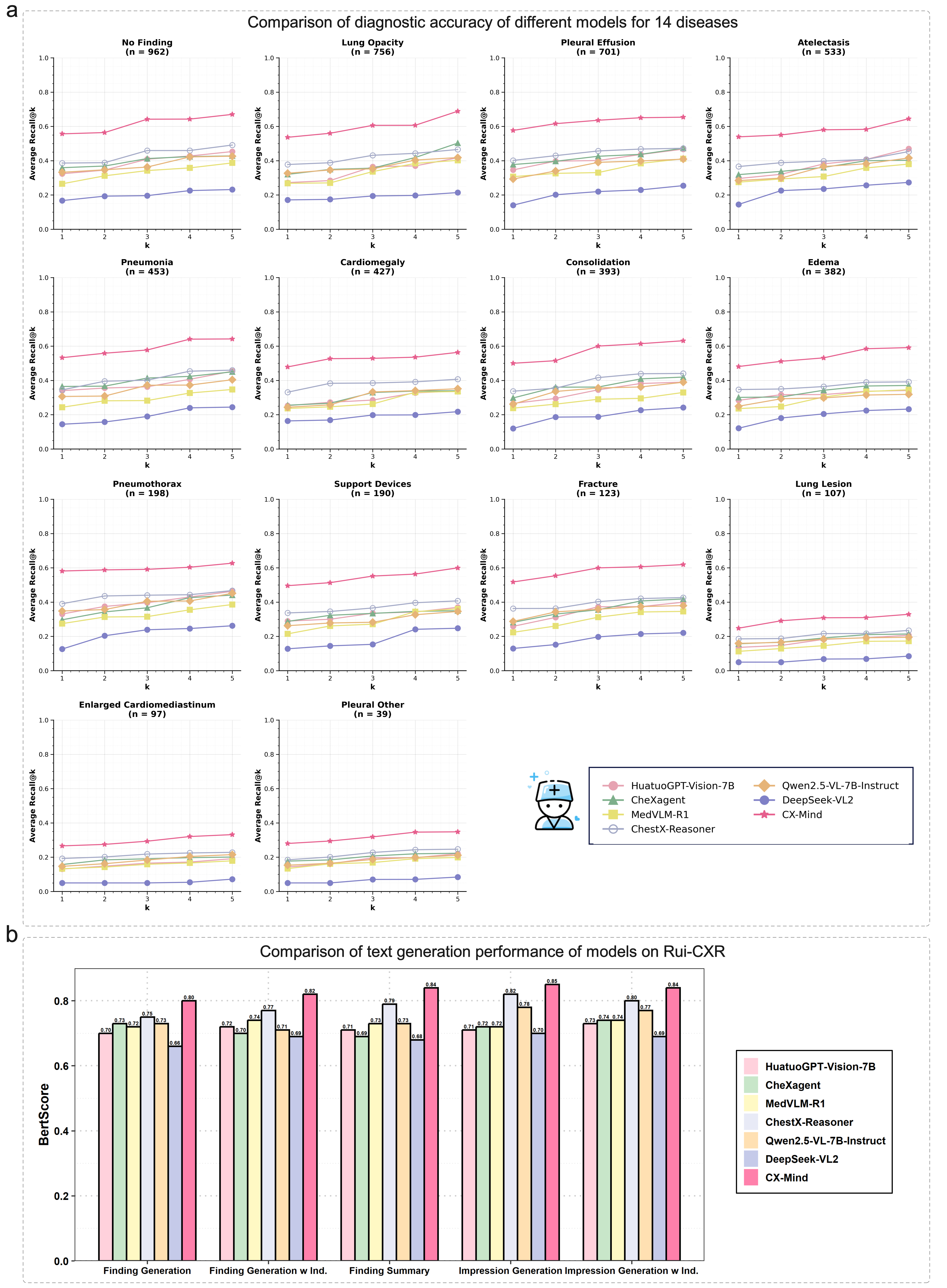} 
    \caption{\textbf{Evaluation of different models on Rui‑CXR.} (a) Diagnostic performance comparison of different models on 14 diseases. (b) Comparison of text generation performance of different models on Rui‑CXR.
}
    \label{fig:Clinical Evaluation} 
\end{figure}

To comprehensively demonstrate \textbf{CX‑Mind}'s outstanding performance, we conducted an extensive cross‑center clinical evaluation. Specifically, the Rui‑CXR dataset originates from 4,031 chest X‑ray image reports collected at the Orthopedics Department of Ruijin Hospital, Shanghai Jiao Tong University, between 2018 and 2023. To protect patient privacy, all images and medical records were rigorously de‑identified and approved by the hospital ethics committee. To prevent data leakage, all comparative models, including CX‑Mind, were evaluated via local deployment.

\subsubsection{Rui‑CXR Dataset Curation}
The Rui‑CXR dataset initially included chest radiographs and accompanying text reports for 80,648 patients acquired by the standard posterior–anterior (PA) chest X‑ray system at the Orthopedics Department of Ruijin Hospital, Shanghai Jiao Tong University, from 2018–2023. To align with the publicly available MIMIC‑CXR dataset, this study focused on 14 common chest diseases (e.g., lung opacity, pleural effusion, Atelectasis, etc.) featured in MIMIC‑CXR. We adopted a three‑stage data filtering strategy: first, we performed an image quality screening on the original 80,648 images, excluding 12,292 samples with severe artifacts or non‑standard PA views. Next, we applied automated keyword matching on the remaining images and reports to select 8,059 candidate images containing descriptions of the target lesions. Based on report completeness checks and diagnostic label consistency verification, we further filtered out 4,028 samples with overly brief reports or mismatched labels, resulting in a high‑quality, fully de‑identified test set of 4,031 images. Since patients may present multiple concurrent diseases, the total count of disease annotations exceeds 4,031.

Based on these 4,031 high‑quality images, we constructed two real‑world clinical evaluation tasks: (1) a disease diagnosis task based on recall@k, where we prompt the model to output $k$ candidate diagnostic labels and quantify the model’s coverage by computing the proportion of true labels in the $k$ predictions. (2) an open‑ended report generation task, in which the model, given a single chest X‑ray input, generates a radiology report conforming to clinical writing standards, and its accuracy and completeness are assessed using quantitative metrics.

\subsubsection{Evaluation Result}
 As shown in Figure \ref{fig:Clinical Evaluation}(a), CX‑Mind demonstrates remarkable diagnostic capabilities on Rui‑CXR: its top‑1 diagnostic accuracy for lung opacity reaches 0.53, significantly outperforming comparative models by 0.20 on average, and achieves the best performance among common diseases such as pleural effusion and Atelectasis, fully validating its high coverage in multi‑candidate diagnostic scenarios. Accuracy declines for conditions such as lung lesion and pleural other, indicating areas for clinical improvement.

As illustrated in Figure \ref{fig:Clinical Evaluation}(b), CX‑Mind secures the highest scores across all five text generation tasks: standard Finding Generation BERTScore 0.80 (w/ Ind. 0.82), averaging a 5\% gain over the second‑best models. These results convincingly demonstrate \textbf{CX‑Mind}'s exceptional ability in fusing chest X‑ray imaging with clinical text for reasoning and high‑quality report generation.

\subsubsection{Expert Assessment}

\begin{figure}[tbp]
    \centering 
    \includegraphics[width=0.9\textwidth]{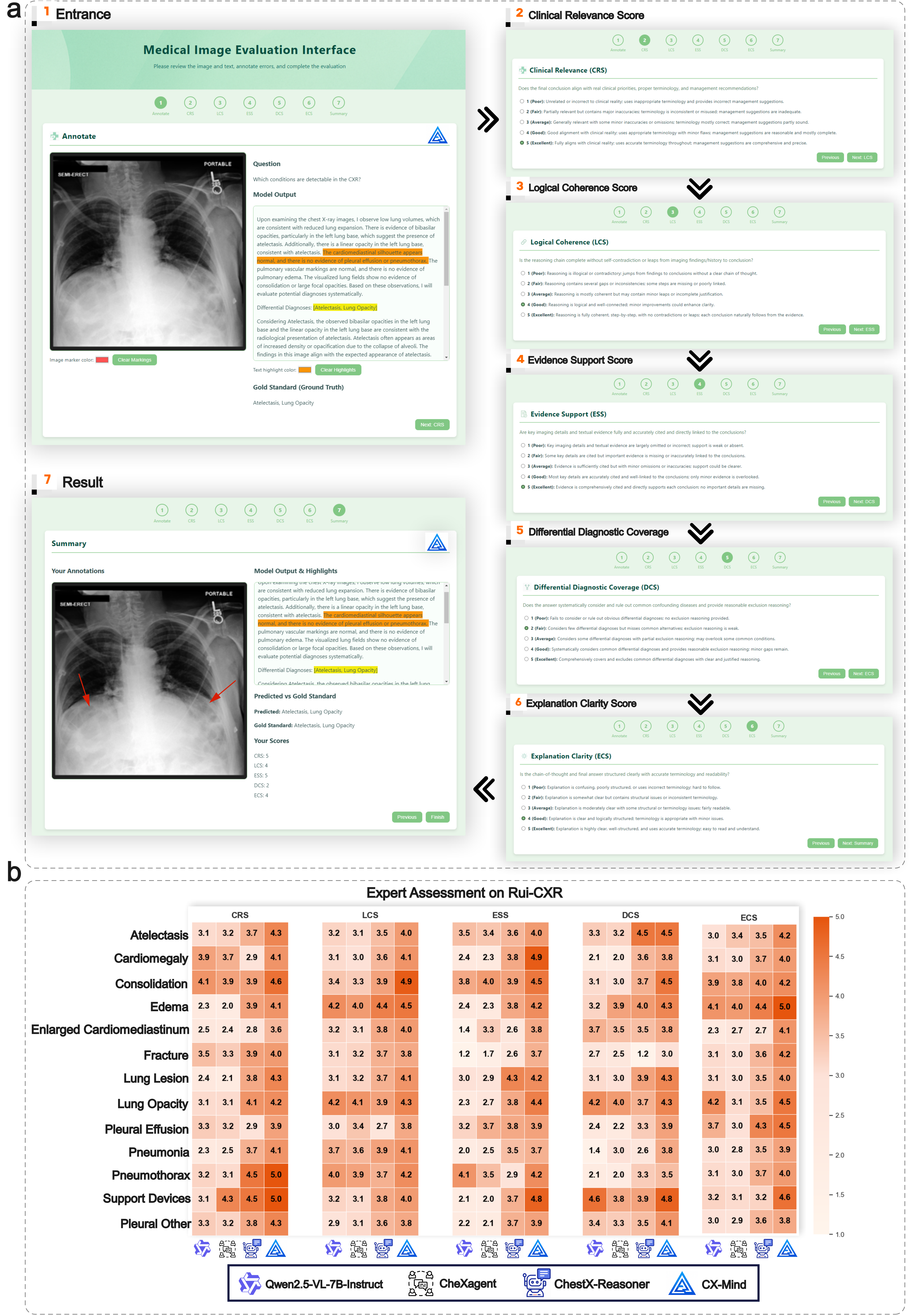} 
\caption{\textbf{Expert evaluation of CX‑Mind and baselines by multi‑center clinicians.} 
(a) Physician scoring workflow interface: the system randomly presents a de‑identified chest radiograph and an anonymized model output. Following unified guidelines, physicians annotate the image and text, then rate CRS, LCS, ESS, DCS, and ECS in sequence. 
(b) Five‑dimension expert‑score heatmap on 13 thoracic diseases from Rui‑CXR. CX‑Mind attains the highest average score on all dimensions.}

    \label{fig:Expert Assessment} 
\end{figure}

We invited internists and surgeons from multiple medical centers to conduct a subjective evaluation of 390 chest radiographs drawn from the Rui‑CXR dataset, covering 13 common thoracic diseases (excluding the "No Finding" category, 30 cases per disease). The diagnostic outputs of Qwen2.5‑VL‑7B‑Instruct, CheXagent, ChestX‑Reasoner, and CX‑Mind were compared. All cases were fully de‑identified, and both model names and any potentially identifying information were masked. The images and model responses were randomly shuffled, yielding a total of 1,560 independent evaluation samples.

For each case, physicians scored the model answer on five dimensions:

\begin{itemize}
    \item CRS (Clinical Relevance Score): whether the conclusion matches clinical priorities, terminology conventions, and management recommendations.
    \item LCS (Logical Coherence Score): whether the reasoning chain is complete and self‑consistent, free of hallucinations and logical jumps.
    \item ESS (Evidence Support Score): whether key image or textual evidence is sufficiently and accurately cited and directly supports the conclusion.
    \item DCS (Differential Diagnostic Coverage): whether common differential diagnoses are systematically considered and reasonably ruled out.
    \item ECS (Explanation Clarity Score): clarity of structure, terminology accuracy, and readability of both the reasoning process and the final answer.
\end{itemize}

To facilitate efficient and convenient assessment, we developed a user‑friendly web interface in Figure \ref{fig:Expert Assessment}(a). The system randomly presents a de‑identified chest radiograph and an anonymous model output. After reading unified guidelines, physicians freely annotate the image (marking suspicious findings) and the text (highlighting inappropriate statements), compare them with the displayed ground‑truth answer, and then complete the above five ratings on a 1–5 scale (1 = poor, 5 = excellent). Approximately 10\% of the cases are re‑sampled to check intra‑rater consistency.

CX‑Mind consistently achieved the highest scores across CRS, LCS, ESS, DCS, and ECS, demonstrating comprehensive superiority from evidence citation to reasoning and final wording. ChestX‑Reasoner ranked at an intermediate level (around 3.5 on average). Its logic and evidence were acceptable, but differential coverage and clarity lagged behind. CheXagent and Qwen2.5‑VL‑7B‑Instruct received lower average scores. Notably, clinicians strongly preferred \textbf{CX‑Mind}'s interleaved reasoning paradigm: they could directly inspect the thought process, rapidly judge its soundness, and intervene promptly when necessary, thereby shortening the feedback loop and enhancing trust and adoption of the model’s output.

\subsection{Qualitative analysis}

\begin{figure}[ht]
    \centering 
    \includegraphics[width=0.9\textwidth]{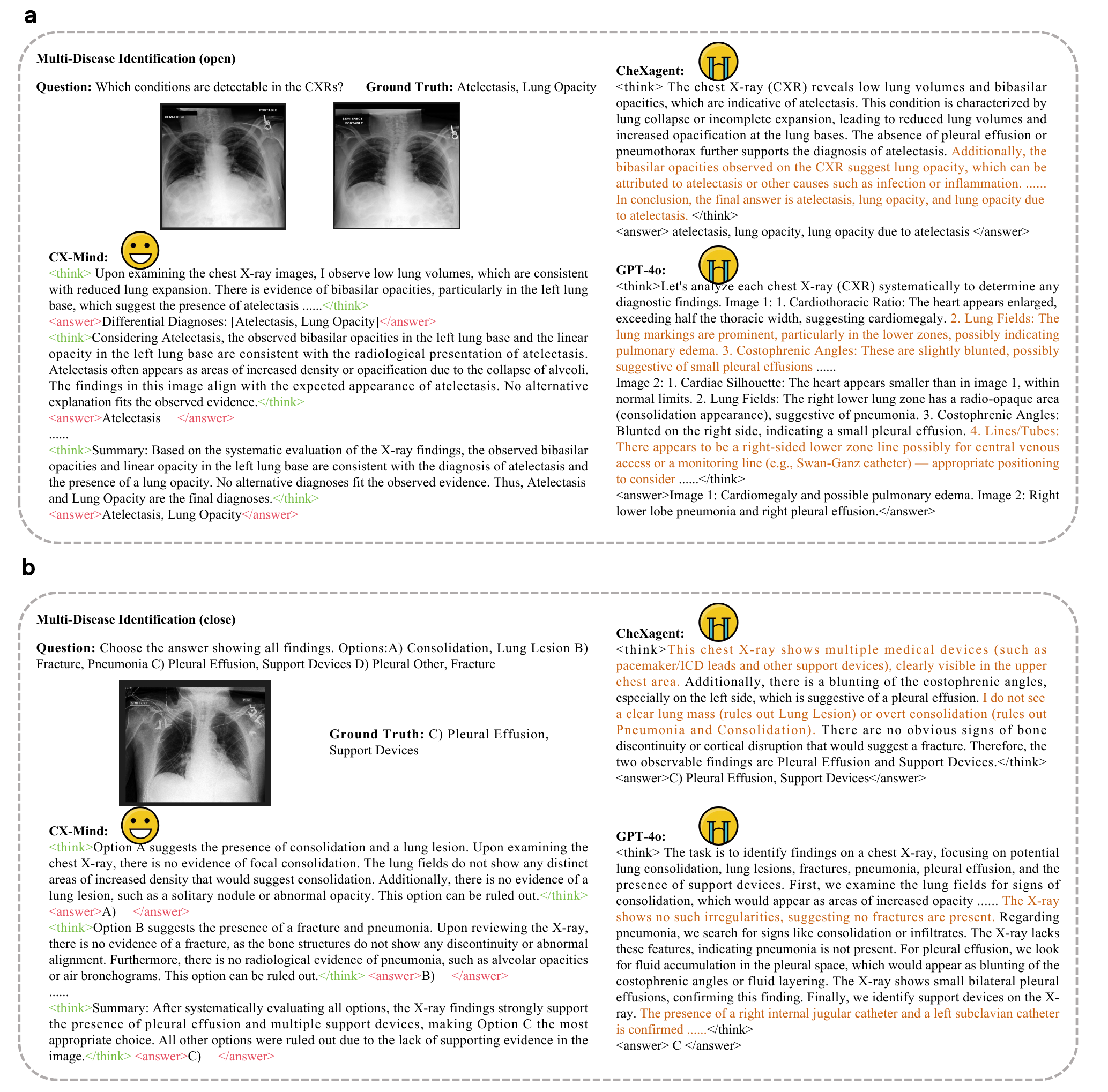} 
\caption{\textbf{Comparative examples of diagnostic responses from GPT-4o, CheXagent and CX-Mind.} (a) Open-ended diagnostic question results. (b) Close-ended diagnostic question results.}

    \label{case} 
\end{figure}

Figure \ref{case} presents a visual comparison of \textbf{CX-Mind}’s performance in the co-morbidity identification task against the closed-source model GPT-4o and the comparable model CheXagent. The content between green and pink tags represents \textbf{CX-Mind}’s reasoning process and intermediate answers (final answer), respectively, while orange annotations highlight information that is inconsistent with medical facts or deviates from clinical observations. The figure clearly demonstrates that both large-scale closed-source models and downstream foundation models fine-tuned on specific X-ray corpora struggle to avoid hallucinations. Such erroneous or fabricated information hinders the practical application of MLLMs in clinical settings. Specifically, in Figure \ref{case}(a), CheXagent attributes lung opacity to potential infection or inflammation, yet the X-ray images lack supporting evidence for these causes (e.g., no typical signs of infection or inflammation, such as pulmonary consolidation). This results in a diagnosis lacking specificity, as it fails to systematically rule out conditions based on available imaging evidence. Similarly, GPT-4o’s reasoning process suggests that increased lung markings may be associated with pulmonary edema. However, this observation alone is insufficient to confirm pulmonary edema, as other conditions (e.g., chronic lung disease) may present similar findings. This process lacks further differential diagnosis to enhance the reliability of the results. In contrast, \textbf{CX-Mind} adheres to our predefined reasoning paradigm, producing a meticulous and accurate interleaved reasoning process. This improves its interactivity with clinicians and enhances its potential for deployment in real-world diagnostic scenarios.

\section{Conclusion \& Discussion}

We have presented CX‑Mind, the first generative chest X-ray diagnostic model that integrates an interleaved reasoning paradigm with structured process rewards within a curriculum learning reinforcement learning framework, achieving a balance between clinical performance and interpretability.
 The training process follows a staged curriculum in which the model first acquires foundational knowledge via SFT, then transitions into RL with GRPO, beginning with close‑ended tasks to establish core reasoning skills before advancing to open‑ended diagnostic queries. By interleaving concise thinking and answering steps, CX‑Mind mirrors the diagnostic workflow of radiologists, allowing each intermediate conclusion to be visible and verifiable. This design not only improves transparency but also enables clinicians to intervene if a reasoning path deviates, thereby reducing the risk of compounding errors. The process reward mechanism, which grants additional feedback only when both format and accuracy criteria are met, encourages the model to refine its reasoning without overwhelming it with premature supervision.

Extensive evaluations demonstrate that CX‑Mind outperforms a broad spectrum of SOTA baselines across disease identification, report generation, and spatiotemporal alignment tasks. In particular, the interleaved outputs yield more focused and succinct chains of thought compared to traditional CoT methods, which often produce verbose and redundant reasoning. Our ablation studies further confirm that the synergy between interleaved reasoning, curriculum RL, and process‑based rewards is critical: without this combined framework, gains in final accuracy come at the expense of coherent intermediate steps, whereas CX‑Mind achieves consistent improvements in both performance metrics and clinician trust.

Despite these advances, real‑world deployment will require careful consideration of several factors. The reliance on publicly available datasets may not fully capture the heterogeneity of clinical imaging protocols or rare pathologies encountered in practice. Moreover, process rewards depend on ground‑truth annotations that might not reflect every nuance of expert radiological judgment. Computational demands for generating and evaluating interleaved reasoning sequences also present challenges for integration into time‑sensitive clinical workflows.

Future work should explore adaptive reward schedules that respond to task complexity and incorporate human feedback loops for on‑the‑fly calibration. Extending the interleaved paradigm and curriculum RL framework to additional imaging modalities and integrating patient history or laboratory findings could further enhance clinical utility. By grounding model outputs in structured, inspectable steps and progressing through a carefully designed curriculum, CX‑Mind paves the way for AI systems that collaborate seamlessly with healthcare professionals, ultimately improving diagnostic accuracy.

\section*{CRediT authorship contribution statement}

\noindent \textbf{Wenjie Li}: Conceptualization, Methodology, Investigation, Formal analysis, Visualization, Writing - original draft, Writing - review \& editing. \textbf{Yujie Zhang}: Software, Data curation, Formal analysis, Writing – review \& editing. \textbf{Haoran Sun}: Methodology, Validation, Software, Formal analysis. \textbf{Yueqi Li}: Investigation, Data curation, Validation. \textbf{Fanrui Zhang}: Software, Visualization. \textbf{Mengzhe Xu}: Formal analysis, Data Curation. \textbf{Victoria Borja Clausich}: Validation. \textbf{Sade Mellin}: Validation. \textbf{Renhao Yang}: Validation. \textbf{Chenrun Wang}: Formal analysis. \textbf{Jethro Zih\textminus Shuo Wang}: Validation. \textbf{Shiyi Yao}: Validation. \textbf{Gen Li}: Validation. \textbf{Yidong Xu}: Validation. \textbf{Hanyu Wang}: Validation. \textbf{Yilin Huang}: Validation. \textbf{Angela Lin Wang}: Validation. \textbf{Chen Shi}: Validation. \textbf{Yin Zhang}: Validation. \textbf{Jianan Guo}: Validation. \textbf{Luqi Yang}: Validation. \textbf{Renxuan Li}: Validation. \textbf{Yang Xu}: Validation. \textbf{Jiawei Liu}: Methodology. \textbf{Lei Liu}: Writing - Review \& Editing. \textbf{Yao Zhang}: Writing - Review \& Editing. \textbf{Carlos Gutiérrez SanRomán}: Validation, Supervision, Writing - Review \& Editing. \textbf{Lei Wang}: Conceptualization, Resources,  Supervision, Project administration, Funding acquisition, Writing – review \& editing.\smallskip

\section*{Declaration of competing interest}
The authors declare that they have no known competing financial interests or personal relationships that could have appeared to
influence the work reported in this paper.

\section*{Acknowledgments}
The work was supported by Key Research and Development Plan of the Ministry of Science and Technology (2023YFC2410705), the National Natural Science Foundation of China (32171317 and 82472419) and Shanghai Municipal Science and Technology Commission Project (25ZR1402348). We gratefully acknowledge the Shanghai Key Laboratory of Intelligent Information Processing at the School of Computer Science, Fudan University, and Professor Bo Yan for their invaluable support.

\section*{Data Availability}
Public Data, the preprocess code will be  available on request. Private datasets cannot be made public due to concerns involving patient privacy.

{
\bibliographystyle{elsarticle-num} 
\bibliography{ref}           
}

\end{document}